\documentclass{article}

\RequirePackage{natbib}

\usepackage{lmodern}
\usepackage{breakcites}

\usepackage[utf8]{inputenc} 
\usepackage[T1]{fontenc}    
\usepackage{url}            
\usepackage{booktabs}       
\usepackage{amsfonts}       
\usepackage{nicefrac}       
\usepackage{microtype}      
\usepackage{graphicx}
\usepackage{subcaption}
\usepackage{amssymb}

\usepackage{amsmath}
\usepackage{cleveref}
\usepackage{chngcntr}

\usepackage{hyperref}

\usepackage[accepted]{icml2019}

\icmltitlerunning{
What is the Effect of Importance Weighting
in Deep Learning?
}

\date{}
\usepackage{color,soul}

\begin{document}

\twocolumn[
\icmltitle{
What is the Effect of Importance Weighting
in Deep Learning?
}



\icmlsetsymbol{equal}{*}

\begin{icmlauthorlist}
\icmlauthor{Jonathon Byrd}{cmu}
\icmlauthor{Zachary C. Lipton}{cmu}
\end{icmlauthorlist}

\icmlaffiliation{cmu}{Carnegie Mellon University}

\icmlcorrespondingauthor{Jonathon Byrd}{jabyrd@cmu.edu}
\icmlcorrespondingauthor{Zachary C. Lipton}{zlipton@cmu.edu}

\icmlkeywords{Machine Learning, ICML, Importance Weighting, Importance Sampling, Deep Learning, Neural Networks, Weighted Risk Minimization}

\vskip 0.3in
]


\printAffiliationsAndNotice{}  
\begin{abstract}
Importance-weighted risk minimization is a key ingredient 
in many machine learning algorithms
for causal inference, domain adaptation, class imbalance, and off-policy reinforcement learning.
While the effect of importance weighting is well-characterized
for low-capacity misspecified models,
little is known about 
how it impacts over-parameterized, deep neural networks.
Inspired by recent theoretical results
showing that on (linearly) separable data,
deep linear networks optimized by SGD
learn weight-agnostic solutions,
we ask, \emph{for realistic deep networks, 
for which many practical datasets are separable, 
what is the effect of importance weighting?}
We present the surprising finding
that while importance weighting impacts deep nets early in training, 
so long as the nets are able to separate the training data,
its effect diminishes over successive epochs.
Moreover, while L2 regularization and batch normalization (but not dropout), 
restore some of the impact of importance weighting,
they express the effect via (seemingly) the wrong abstraction:
why should practitioners tweak the L2 regularization,
and by how much, to produce the correct weighting effect?
We experimentally confirm these findings 
across a range of architectures and datasets. 
\end{abstract}

\section{Introduction}
\label{sec:intro}
Importance sampling is a fundamental tool 
in statistics and machine learning often used
when we want to estimate a quantity on some \emph{target} distribution, 
but can only sample from a different \emph{source} distribution
\citep{horvitz1952generalization, kahn1953methods, rubinstein2016simulation, koller2009probabilistic}. 
Concretely, given $n$ samples $x_1, ..., x_n \sim p(x)$,
and the task of estimating some function of the data, 
say $f(x)$, under the target distribution $E_q[f(x)]$,
importance sampling produces an unbiased estimate 
by weighting each sample $x$ 
according to the likelihood ratio $q(x)/p(x)$:
\begin{align*}
\mathbb{E}_p\left[ \frac{q(x)}{p(x)} f(x)\right]
&= \int_{x} f(x) \frac{q(x)}{p(x)} p(x) dx \\
= \int_{x} f(x) q(x) dx &= \mathbb{E}_q\left[f(x)\right]
\end{align*}
Machine learning practitioners commonly exploit this idea in two ways:
(i) by re-sampling to correct for the discrepancy in likelihood or 
(ii) by weighting examples according to the likelihood ratio 
\citep{rubinstein2016simulation, shimodaira2000improving, koller2009probabilistic}. 
For this reason, among others, weighted risk minimization is a standard tool
that emerges in a wide variety of machine learning tasks.

In domain adaptation, when source and target data share support,
practitioners commonly adjust for distribution shift 
by estimating likelihood ratios. This is done either as a function of the input $x$: $q(x)/p(x)$ 
(in the case of covariate shift)
or of the label $y$: $q(y)/p(y)$ (in the case of label shift), 
training a \emph{corrected} classifier with \emph{importance-weighted empirical risk minimization (IW-ERM)}  
\citep{shimodaira2000improving, gretton2009covariate, lipton2018detecting}.
One related use of IW-ERM is to correct for sampling bias 
in active learning \citep{Beygelzimer2009importance, settles2010active}.
The technique is also frequently employed in
off-policy reinforcement learning 
\citep{precup2000eligibility, mahmood2014weighted, swaminathan2015counterfactual},
where we desire to learn a new policy given offline samples collected 
from a preexisting policy. 
Weighted loss functions also arise in a number of other contexts,
including label noise and crowdsourcing.



\subsection{Deep learning and weighted risk minimization}

When our hypothesis class consists of low-capacity models
that are misspecified,
importance weighting has well-known benefits.
Consider the simple case of fitting a linear model 
to data generated by a higher-order polynomial.
For a reasonably large training set,
our model must make errors somewhere. 
By altering the relative contribution 
of mistakes on various training points
to our loss function,
importance weights typically lead us to fit a different model.

As deep learning has come to dominate a broad set of prediction tasks,
importance-weighted risk minimization
has remained a standard technique.
\cite{shalit2017estimating} employ neural networks
to estimate individual treatment effects by
weighting their loss function to compensate 
for differences in treatment group size. 
In a work on deep learning and crowdsourcing, 
\citet{khetan2017learning} proposed a weighted loss 
as part of an iterative scheme for jointly estimating worker quality 
and learning a classifier from noisy data. 
In recent work on label shift,
\citet{lipton2018detecting, azizzadenesheli2019regularized},
propose adapting deep networks with importance-weighted risk minimization.

Applications of importance weighting also abound
in deep reinforcement learning.
For example, \citet{joachims2018deep} use the technique
to learn from logged contextual bandit feedback.
Other applications include deep imitation learning \citep{murali2016tsc}.
In one paper, \citet{schaul2015prioritized},
employ a weighted sampling to choose experiences 
from the replay buffer for performing TD updates.
These weights are not based on likelihood ratios, 
but are chosen heuristically
to be proportional to the Bellman errors.

Despite the popularity of importance sampling
in combination with deep neural networks,
how and when it works remain open questions.
Unlike linear models, deep neural networks are generally over-parameterized,
capable of fitting training datasets to perfect accuracy \cite{zhang2017understanding}.
Moreover, it is now recognized that for many tasks deep neural networks 
continue to improve generalization error
past the point of achieving zero training error \cite{soudry2017implicit}.
Thus they are not only \emph{capable} of separating the training set (given enough epochs)
but actually are trained to do so in common practice.
Since neural networks are capable of shattering the training set (and often do),
it is not clear that any trade-offs must be made
among classifying each of the training points. 
Thus any effects of importance weighting
depend crucially on how they impact the dynamics of optimization,
an actively-studied but still poorly-understood topic. 

\subsection{Salient findings}
In this paper, we investigate the effects of importance weighting
in deep learning across a variety of architectures, tasks, and data sets.
We present the surprising result 
that importance weighting may or may not have any effect
in the context of deep learning, depending on particular choices regarding 
early stopping, regularization, and batch normalization.
Our experiments focus on classification problems:
we apply class-conditioned weights of various strengths,
evaluating the impact of the weights
on the learned decision boundaries.
We consider the effect of weighting 
on both how training and test points are classified,
and also the effect on off-manifold data points and random noise.
Our experiments address both the classification of CIFAR-10 images
and paraphrase detection using data from the 
Microsoft Research Paraphrase Corpus (MRPC).
We also build intuition by considering 2D synthetic datasets 
for which we can visualize decision boundaries.

Across tasks, architectures and datasets,
our results confirm that for standard neural networks, 
weighting has a significant effect early in training.
However, 
as training progresses 
the effect dissipates
and for most weight ratios considered (between 256:1 and 1:256)
the effect of importance weighting is indistinguishable 
from unweighted risk minimization after sufficient training epochs. 
While L2 regularization restores some of the impact of importance weighting,
this has the perplexing consequence 
of expressing the \emph{amount} by which importance weights affect the learned model
in terms of a seemingly unrelated quantity---the degree of regularization---prompting the question:
\emph{how does one appropriately choose the L2 regularization 
given importance weights?}
Interestingly, dropout regularization,
which is often used interchangeably with L2 regularization,
does not exhibit any such interaction with importance weighting. 
Batch normalization also appears to interact with importance weights,
although as we will discuss later, the precise mechanism remains unclear.

\subsection{Contributions}

In summary, our contributions are the following:
\begin{enumerate}
    \item We demonstrate the surprising finding 
    that for unregularized neural networks 
    optimized by \emph{stochastic gradient descent (SGD)},
    the impact of importance weighting diminishes over epochs of training.
    \item We show that L2 regularization and batch normalization, 
    (but not dropout) interact with importance weights,
    restoring (some) impact on learned models.
    \item We replicate our results across a variety of networks, 
    tasks, and datasets.
    \item Our results call into question the standard application 
    of importance weighting when applied to deep networks, 
    a finding with practical consequences
    on the fields of causal inference, domain adaptation, 
    and off-policy reinforcement learning. 
\end{enumerate}

\section{Theoretical Motivation}
\label{sec:motivation}
The empirical questions addressed in this paper 
draw inspiration from recent developments 
in the theory of deep learning.
In particular we are motivated by finds of \citet{soudry2017implicit} and \citet{gunasekar2018implicit},
who investigate the decision boundaries learned by neural networks.
Note that although these theoretical analyses 
cover only \emph{shallow linear}, \emph{deep linear}, 
and \emph{deep convolutional linear} neural networks, 
our experiments draw intuition from the results and confirm empirically
that the hypotheses hold on more practical nonlinear networks.

\citet{soudry2017implicit} note that it is common practice
to train neural network classifiers to overfit badly,
and that even past the point (in terms of training epochs)
of achieving zero training error,
although the negative log-likelihood on holdout data begins to increase,
the generalization error often continues to decrease. 
To analyze this phenomenon, they restrict their attention to linear networks 
which are presently more amenable to the available tools of analysis. 

They consider the simple case where the model consists of a linear separator,
the data is linearly separable,
the optimization objective is cross-entropy loss,
and the optimization algorithm is SGD.
Notably, there is no finite minimizer $w^*$ of the objective,
since for any $w$ that separates the data,
an even lower loss could be achieved by scaling up the weights $w$.
Thus the weights themselves do not converge. 
However, noting that the learned decision boundary depends
only on the direction of the weights 
(but not their magnitude), 
\citet{soudry2017implicit} examine what, if anything, 
the \emph{direction} $w_t/||w_t||$ converges to 
(over training iterations of SGD). 
Surprisingly, they conclude that the weights converge in direction 
to the solution of the hard-margin \emph{support vector machine}. 
In short, the proof follows because over epochs of training,
the norm of the weight vector increases,
causing the support vectors to dominate the loss function
(under a set of conditions satisfied by the cross-entropy loss).
Subsequent results confirm that this finding holds 
for deep fully-connected networks of linear units \cite{gunasekar2018implicit},
and that for deep convolutional networks of linear units,
a related result holds \cite{gunasekar2018implicit},
showing implicit bias towards minimizing the $\ell_{2/L}$
bridge penalty in the frequency domain 
of the corresponding single-layer linear predictor.

One interesting ramification of this theoretical result 
is that the hard-margin solution depends 
only on the \emph{location} of data points, 
and thus is unaffected by \emph{oversampling/re-weighting}. 
While \citet{soudry2017implicit} and \citet{gunasekar2018implicit}'s analyses 
only address linear networks and linearly-separable data,
their findings motivate our hypothesis that 
a similar weight-invariance property might hold 
for typical modern deep (nonlinear) neural networks, 
for which many datasets of practical interest are separable \citep{zhang2017understanding}. 

These results also motivate our follow-up questions
concerning the effect of regularization. 
Common regularization methods like L2 regularization penalize 
the large-norm solutions that minimize cross-entropy on separable data. 
Since L2 regularization prevents such large-norm solutions, 
what if anything is the impact of importance weights in this case? 
Moreover, while \textit{dropout} \citep{srivastava2015dropout} 
is often thought of as a regularization method for deep networks, 
it does not penalize large-norm solutions.
Thus we hypothesize that these regularization methods 
would have differential impacts on the solutions found 
by SGD on deep networks in conjunction with importance weighting.

\section{Experiments}
\label{sec:experiments}


We investigate the effects of importance weighting on neural networks 
on two-dimensional toy datasets, the CIFAR-10 image dataset, 
and the Microsoft Research Paraphrase Corpus (MRPC) text dataset. 
Our experiments address the label shift scenario, weighting examples based on their class. 
Specifically, we down-weight the loss contributions of examples from a particular class. 
We also test the combination of regularization and IW-ERM on both CIFAR-10 and a toy dataset. 
For L2 regularization, we set the penalty coefficient as $0.001$, 
and when using dropout on deep networks, 
we set the values of hidden units to $0$ 
during training with probability $\frac{1}{2}$. 


\begin{figure*}[t]
\includegraphics[width=\linewidth]{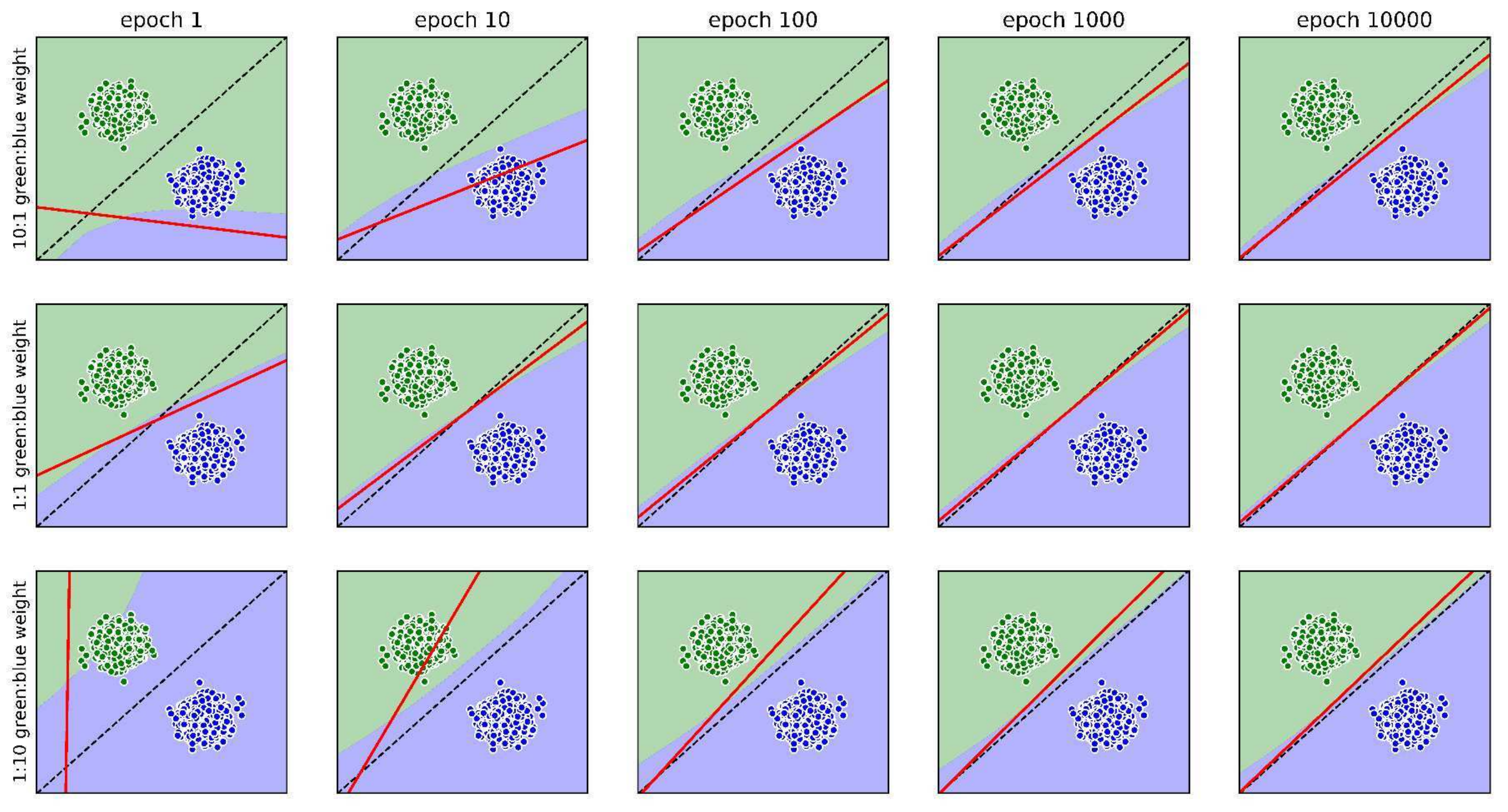}
\caption{Convergence of decision boundaries over epochs of training 
with different importance weights (top to bottom). 
Points are colored according to their true labels, 
with background shading depicting the decision surface 
of an MLP with a single hidden layer of size $64$. 
The red line shows the logistic regression decision boundary. 
The dotted black line shows the max-margin separator.}
\label{fig:2dsynth}
\end{figure*}

\begin{figure*}
\includegraphics[width=\linewidth]{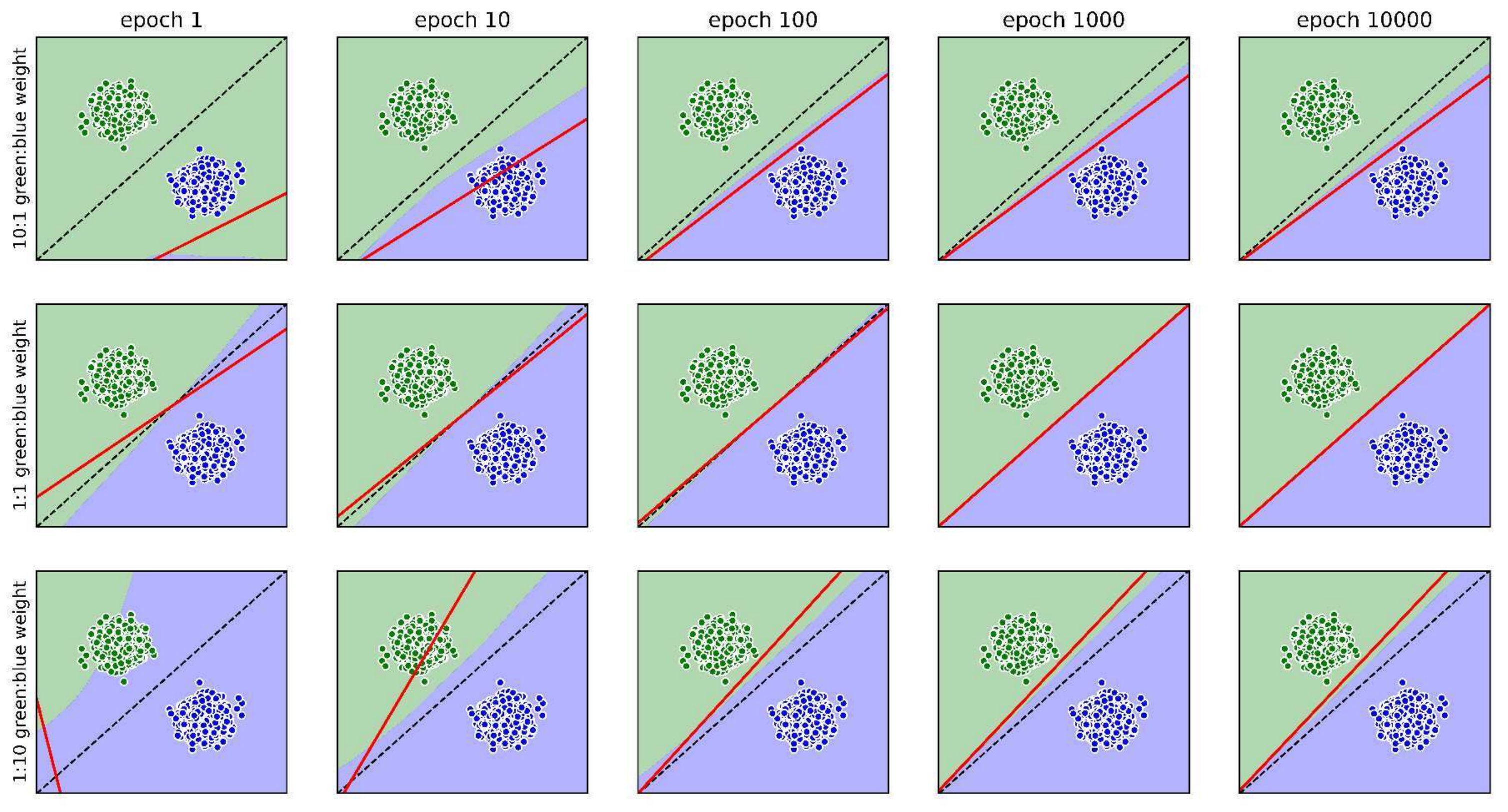}
\caption{Same scenario as Figure ~\ref{fig:2dsynth}, except both logistic regression and MLP are trained with L2 regularization.}
\label{fig:2d_l2reg}
\end{figure*}

\begin{figure*}
\includegraphics[width=\linewidth]{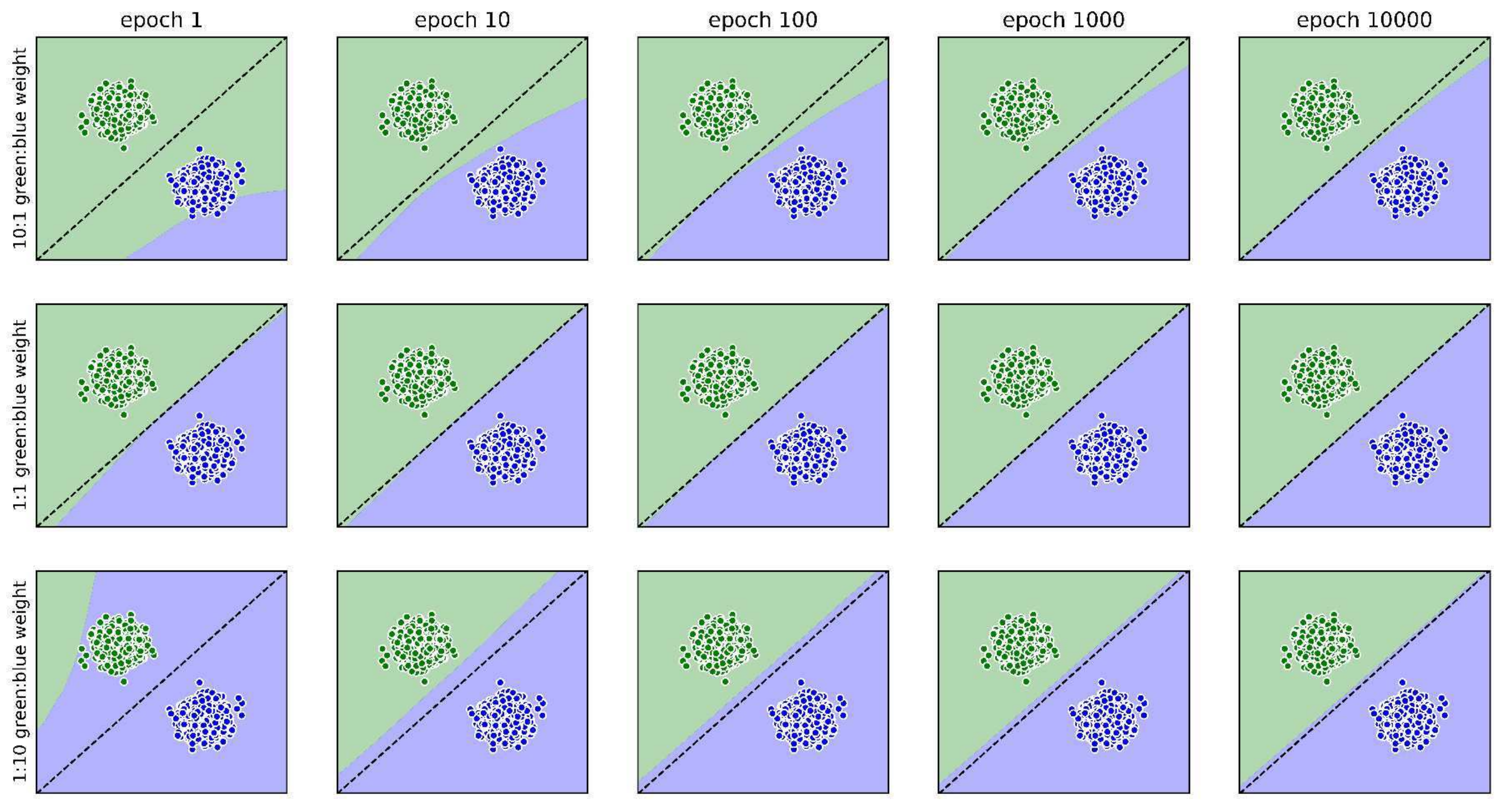}
\caption{Same scenario as Figures ~\ref{fig:2dsynth} and ~\ref{fig:2d_l2reg}, 
except the MLP is trained with dropout (no logistic regression model shown).}
\label{fig:2d_dropout}
\end{figure*}

\paragraph{Synthetic Data} 
In order to visualize decision surfaces, 
we conduct an experiment with a synthetic two-dimensional linearly-separable dataset. 
To form the \emph{positive examples}, we sample $512$ points 
from a 2D truncated normal distribution.
To generate \emph{negative examples}, we rotate and translate 
the positive examples (see Figure ~\ref{fig:2dsynth}).
We train both a logistic regression model (without regularization) 
and a multi-layer perceptron (MLP) using minibatch SGD 
for 10,000 epochs with a batch size of $8$. 
The MLP has a single hidden layer of $64$ hidden units with ReLU activations. 
Both models use a fixed learning rate of $\frac{1}{\sigma_{max}(\mathbf{X})}$, 
where $\sigma_{max}(\mathbf{X})$ is the maximum singular value of the data matrix. 
This learning rate was chosen to match the experiments of \citet{soudry2017implicit}, 
and took a value of $\approx  0.045$ on our dataset. 
Results are shown in Figures ~\ref{fig:2dsynth}, ~\ref{fig:2d_l2reg}, and ~\ref{fig:2d_dropout}. 
We also present results for experiments on a two-dimensional moons dataset 
and a two-dimensional overlapping Gaussian distribution dataset 
that are not linearly-separable (Figures ~\ref{fig:2d_moons} and ~\ref{fig:2d_noise}).
 
%

\begin{figure*}[!t]
    \centering
    \subcaptionbox{CIFAR-10 cat and dog test images.}%
    [.32\textwidth]{\includegraphics[width=.32\textwidth]{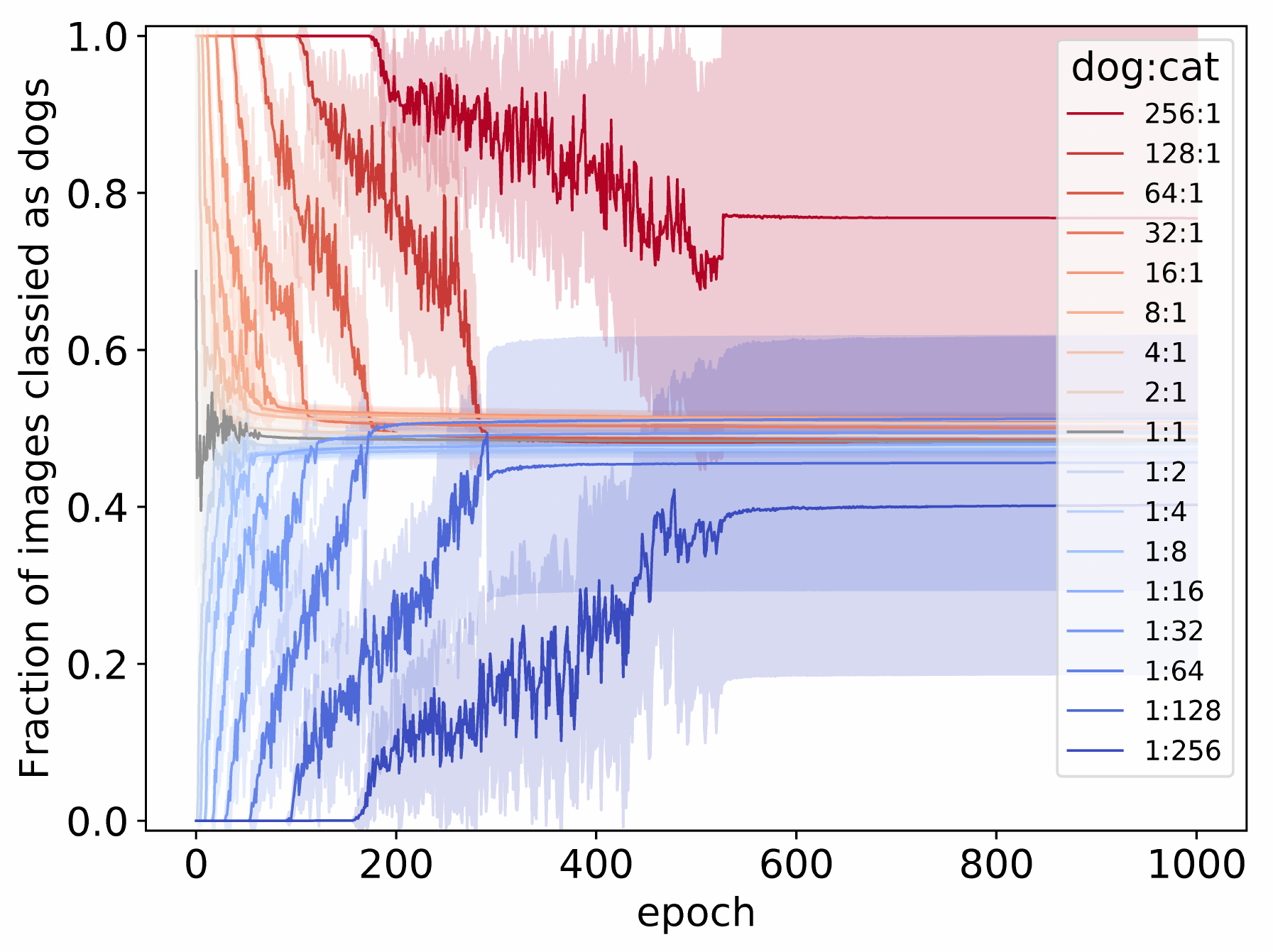}}
    \label{fig:cifar_epochs_catdog}
    \hfill
    \subcaptionbox{CIFAR-10 test images from the other eight classes.}%
    [.32\textwidth]{\includegraphics[width=.32\textwidth]{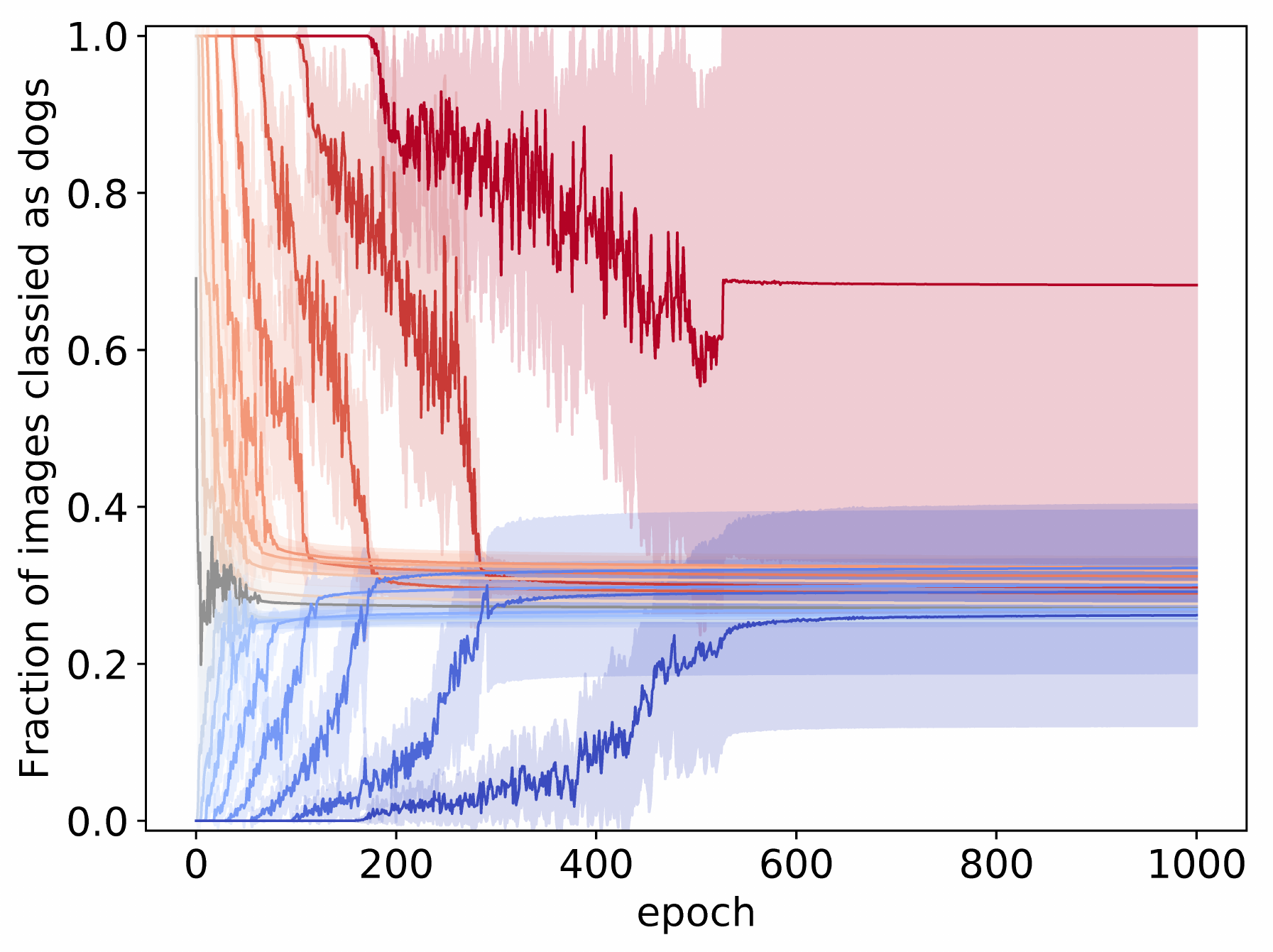}}
    \label{fig:cifar_epochs_other}
    \subcaptionbox{Random images.}%
  [.32\textwidth]{\includegraphics[width=.32\textwidth]{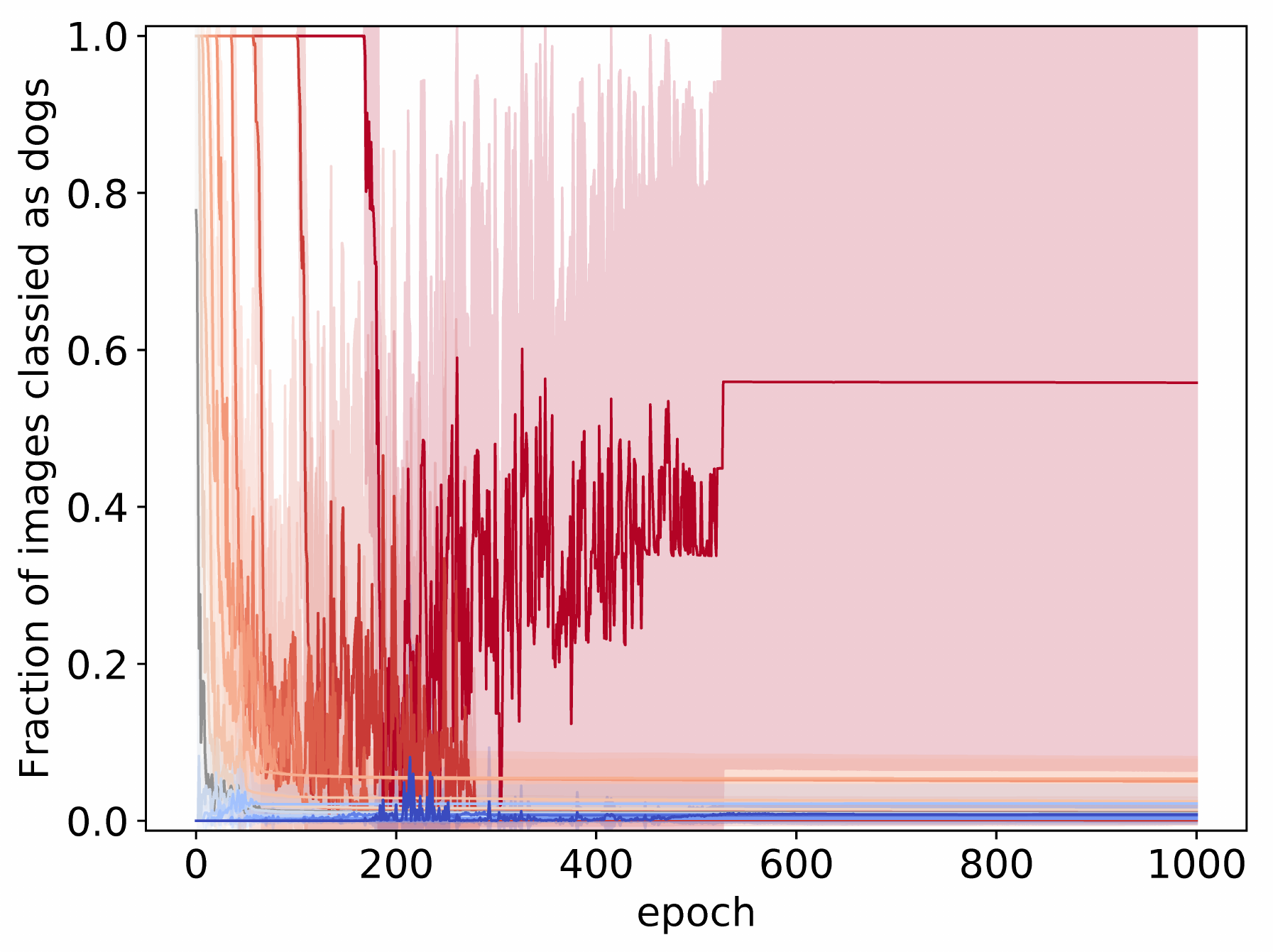}}
    \label{fig:cifar_epochs_random}
    \vskip\baselineskip
    \subcaptionbox{Classification ratios at epoch $1000$.}%
    [.32\textwidth]{\includegraphics[width=.32\textwidth]{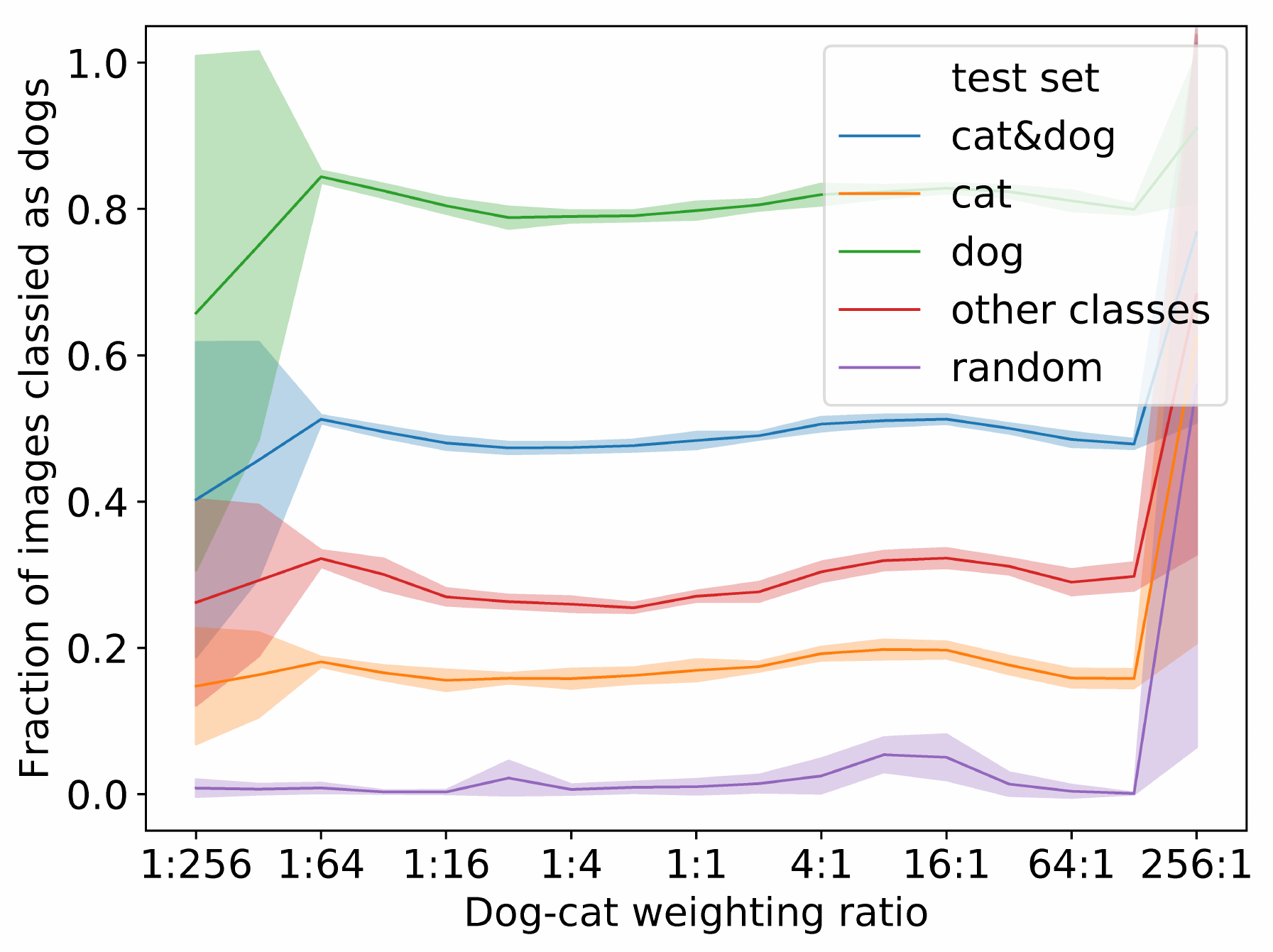}}
    \label{fig:cifar_iw}
    \hfill
    \subcaptionbox{Classification ratios with L2 regularization on weights.}%
    [.32\textwidth]{\includegraphics[width=.32\textwidth]{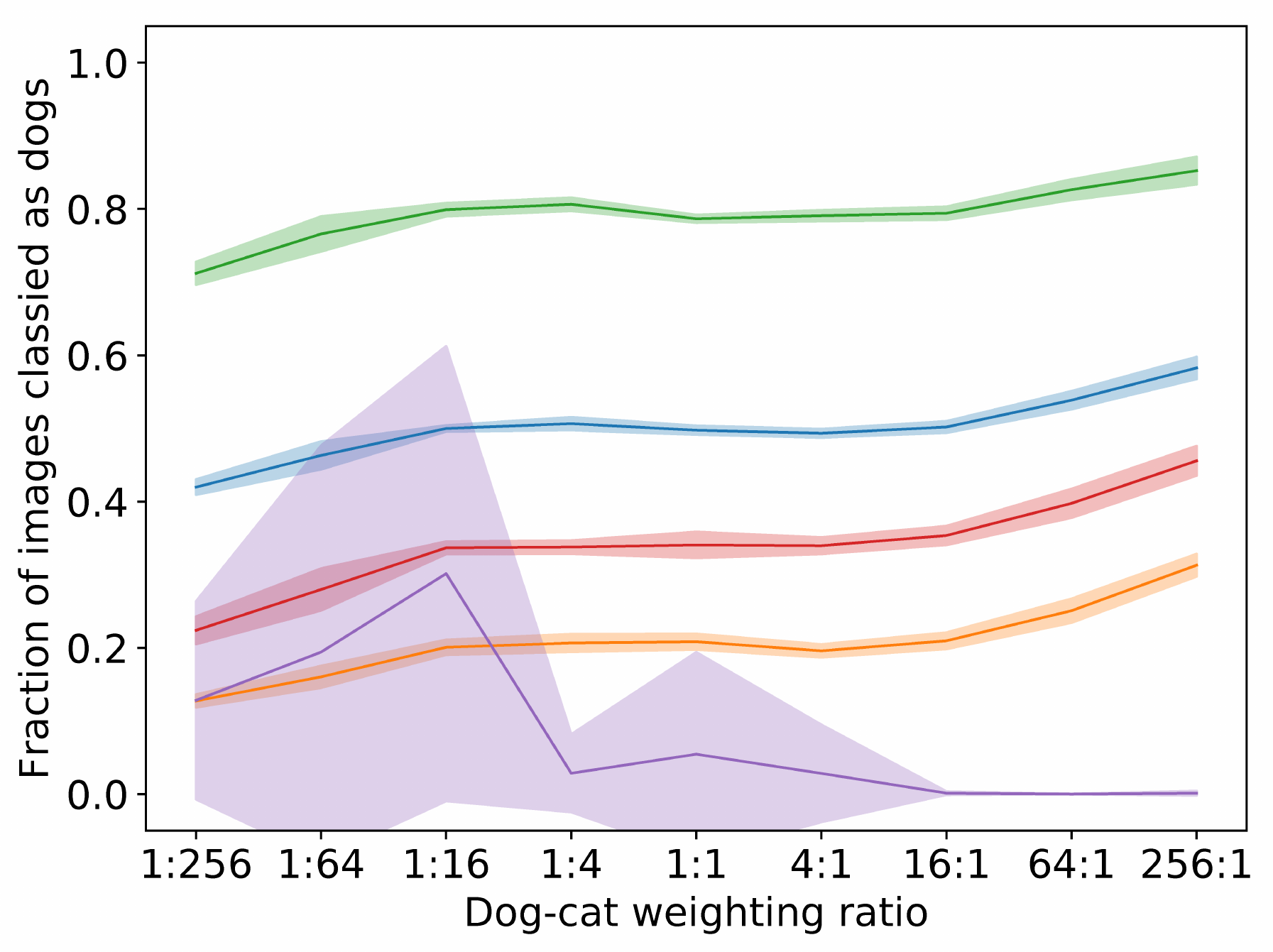}}
    \label{subfig:cifar_l2reg}
    \hfill
    \subcaptionbox{Classification ratios with dropout.}%
    [.32\textwidth]{\includegraphics[width=.32\textwidth]{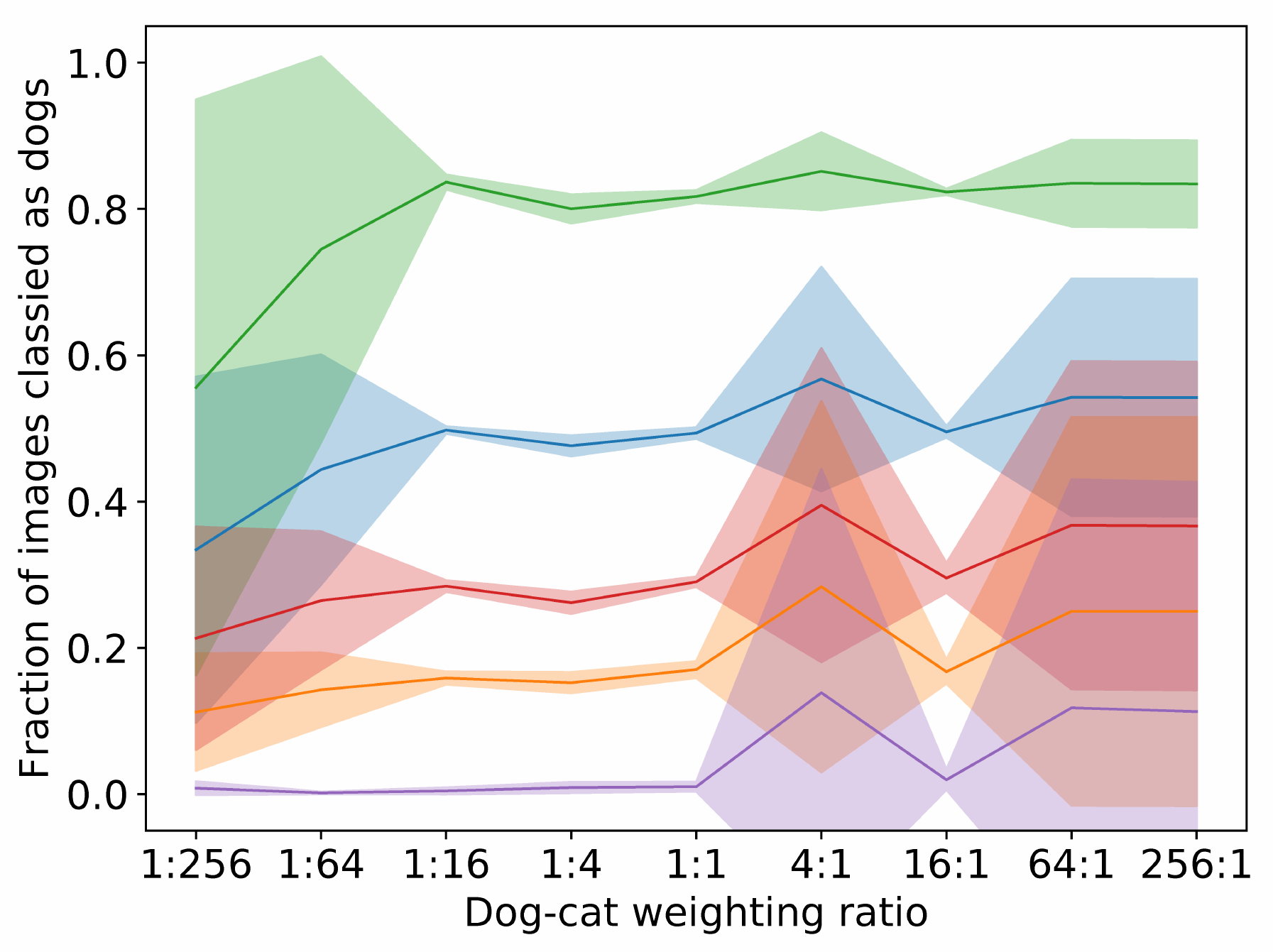}}
    \label{subfig:cifar_dropout}
    \caption
    {(a-c) Relationship between early stopping and importance weighting.
    We plot the fraction of images in the test set (a), other 8 classes (b), 
    and random vectors (c), classified as dogs (y-axis) vs training epochs (x-axis).
    (d-f) Fraction  of examples classified as dogs (y-axis) 
    vs importance weights (x-axis) after 1000 epochs of training. 
    We also show results from models trained with L2 regularization (e) and dropout (f).
    In all plots error bands show standard deviation across nine random initializations, 
    and lines represent means.} 
    \label{fig:cifar}
\end{figure*}

\paragraph{CIFAR-10 Binary Classification} 
We also conduct experiments on the CIFAR-10 dataset 
(see results in Figure ~\ref{fig:cifar}). 
Here, we train a binary classifier on training images 
labeled as cats or dogs ($5000$ per class), 
evaluating on all $10000$ test images from all $10$ classes 
as well as $1000$ random noise images. 
The classifier is a convolutional network 
with the following structure: 
two convolution layers with $64$ $3 \times 3$ filters each and stride $1$, 
followed by a $2 \times 2$ max pooling layer, 
followed by three convolution layers with 
$128$ $3 \times 3$ filters each and stride $1$, 
followed by a second $2 \times 2$ max pooling layer, 
followed by two dense layers with $512$ and $128$ hidden units respectively, 
and finally the binary output layer. 
All hidden layers employ ReLU activation functions. 
The models are trained for $1000$ epochs using minibatch SGD 
with a batch size of $16$ and no momentum.  
All models trained with SGD use a constant learning rate of $0.1$, 
except for the dropout models with no importance weighting 
which used a learning rate of $0.05$ due to weight divergence issues. 
We also ran experiments with the Adam optimizer \citep{kingma2015adam} 
with learning rate $1e-4$, $\beta_1=0.9$, $\beta_2=0.999$, and  $\epsilon=1e-8$ (Figure \ref{fig:adam}).
Experiments were run with importance weights of inverse powers of $2$ 
up to $2^{-8}$ for each class, 
as well as with no importance weighting for unregularized models.
Results are given for importance weights of inverse powers of $4$ for regularized models.
Figure ~\ref{fig:labelnoise} shows results on CIFAR-10 cats and dogs
with label noise in the training data 
where the underlying noisy distributions are not separable, but the finite sample is separable. 
We create samples with label noise by flipping the labels 
of $5\%$ of training examples from each class.
Figure ~\ref{fig:cartruck} shows results from training
the convolutional network model on CIFAR-10 images labeled automobile or truck.

All unregularized classifiers without data sub-sampling or label noise achieve (unweighted) test accuracy 
between $80\%$ and $85\%$. 
In addition to noting the similarity of models across important weightings
both in terms of accuracy and the fraction of examples 
predicted to belong to each category,
we investigated the extent to which the models agreed with each other
on precisely which examples belonged to each class.
We compare (i) the agreement between models with different importance weights 
with (ii) the agreement between models run with different random seeds.
To compute (i), we first compute test set predictions for each importance weighting 
by taking a majority vote over $9$ different random seeds. 
We find that, on average, $82\%$ of the $17$ differently-weighted models 
agree on the label of a given test set example 
($74\%$ agreement on out-of-sample CIFAR images from the other $8$ classes).
For (ii), we calculate the fraction of random initializations 
that agree on each example for each importance weighting. 
Then we average over both examples and importance weightings 
to find that, on average, $78\%$ of random initializations agree on the label of a given test set example ($71$\% agreement on out-of-sample images). 
We note that all models have near-perfect agreement 
on random noise images which are nearly-always classified as cats.
%

We repeat the same CIFAR experiment using the popular 
deep residual network (ResNet) architecture \citep{he2016deep}
consisting of a $5 \times 5$ convolution with $64$ filters 
followed by two residual blocks with $64$ filters, 
then two residual blocks with $128$ filters, 
then two residual blocks with $256$ filters, 
followed by average pooling, a dense layer with $512$ nodes, and finally, the output layer. 
Each residual block consists of two $3 \times 3$ convolution layers. 
The first layer with $128$ filters and the first layer with $512$ filters have stride of $2$. 
All other hyperparameters were left unchanged. 
Figure ~\ref{fig:resnet} shows results 
both with and without batch normalization \citep{ioffe2015batch} 
applied between all convolution layers.

\paragraph{CIFAR-10 Imbalanced}
Importance weighting is commonly used to correct for class imbalance.
To simulate this situation, we train on an imbalanced training set 
created by sub-sampling CIFAR-10 examples from either the cat or dog class 
(Figure \ref{fig:subsamp}). 
We downweight the loss function for the class 
that wasn't sub-sampled by the same factor used to sub-sample the other class.


\begin{figure}
\centering
\includegraphics[width=.47\textwidth]{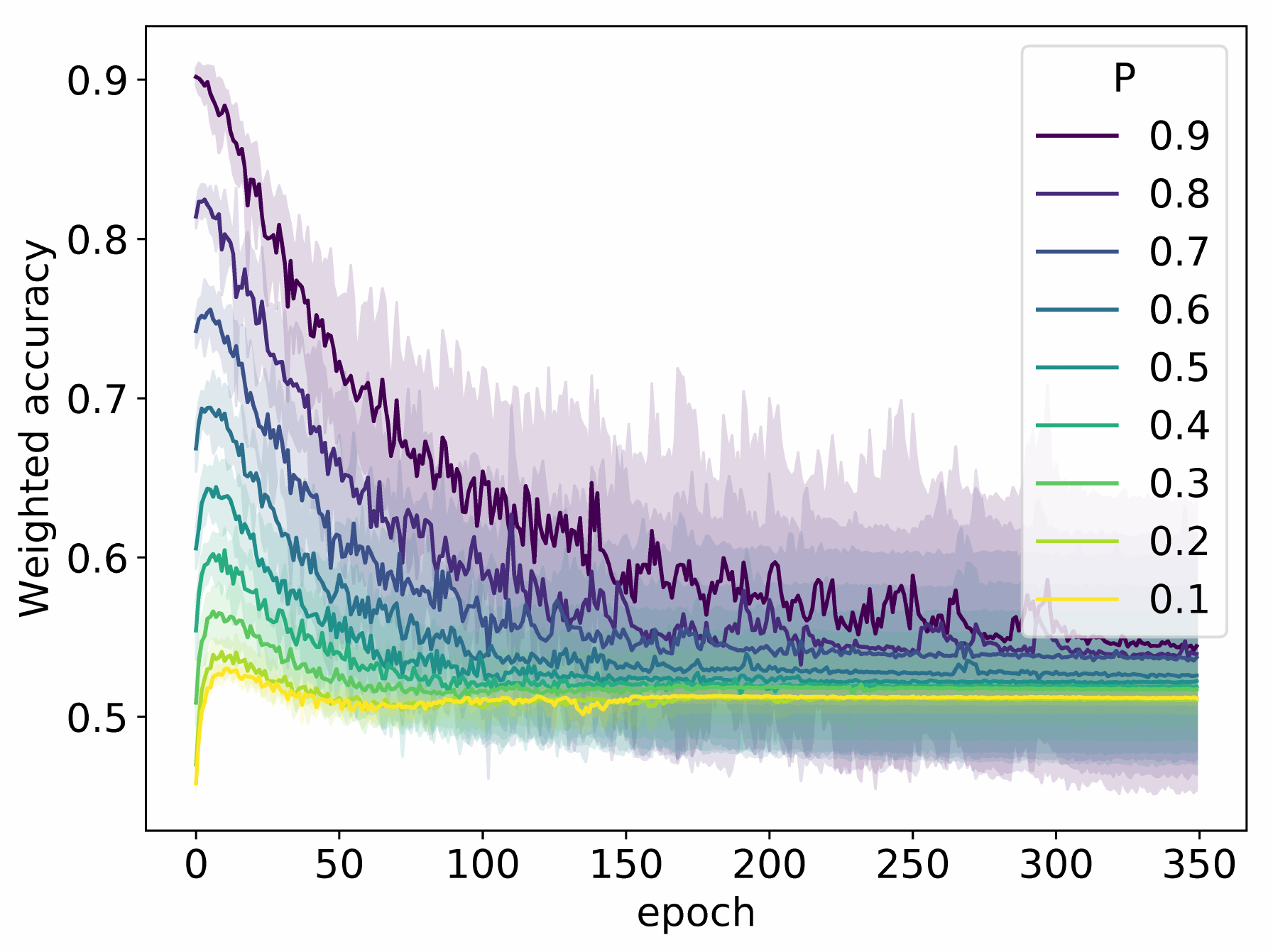}
\caption{
Effect of early stopping on test label shift correction for multiclass CIFAR-10. 
For each value of $\rho \in [0.1\ldots0.9]$, 
we weight one class's contributions 
to the loss function and test accuracy by $\rho$, 
and the other nine classes by $\frac{1 - \rho}{9}$. 
We plot the relationship between the weighted test accuracy (y-axis) 
vs training epochs (x-axis).
Lines and errors show the means and standard deviations over the ten CIFAR-10 classes respectively.}
\label{fig:multiclass}
\end{figure}

\paragraph{CIFAR-10 Multiclass}
In addition to binary classification, we also explore class weighting 
in the multiclass setting using all CIFAR-10 classes (Figure \ref{fig:multiclass}). 
We set up experiments similar to \citet{lipton2018detecting}, 
where we weight the loss function contributions of one class by $\rho \in [0.2, 0.5, 0.9]$, 
and applying a weight of $\frac{1 - \rho}{9}$ to the other nine classes. 
At test time, we apply the same weights to each class's contribution to accuracy 
in order to simulate correcting for label shift in the test set. 
Here, the classifier is a two-layer MLP with $256$ hidden units 
trained using weighted cross-entropy loss with a learning rate of $0.01$.


\begin{figure}
\centering
\includegraphics[width=.47\textwidth]{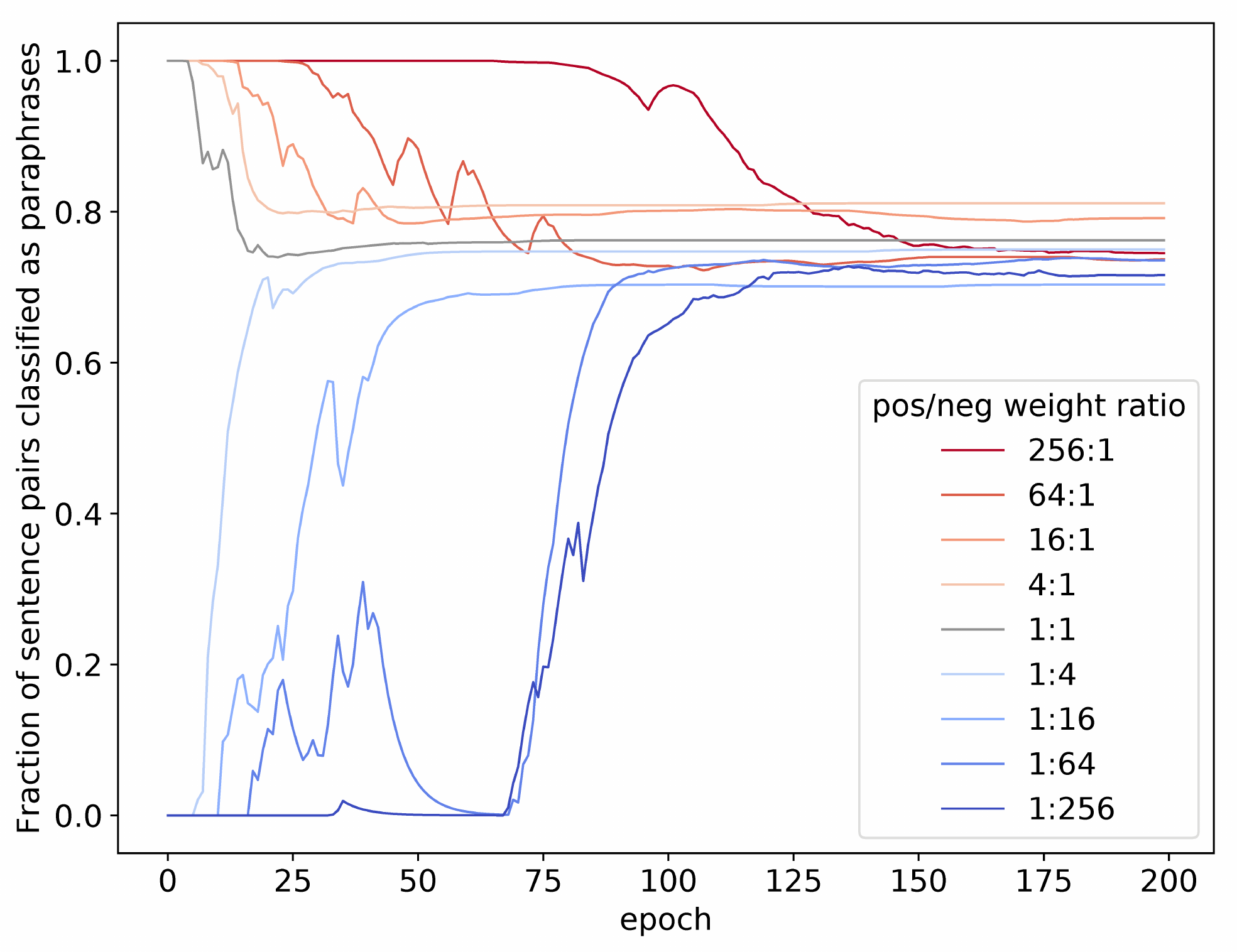}
\caption{Relationship between early stopping and importance weighting on MRPC dataset.  
We plot the fraction of sentence pairs in the test set, 
classified as paraphrases (y-axis) vs training epochs (x-axis) with different importance weights.}
\label{fig:mrpc}
\end{figure}

\paragraph{MRPC} 
To verify that our findings hold in other domains, 
we conduct similar experiments on (sequential) natural language data 
using the Microsoft Research Paraphrase Corpus (MRPC) \citep{dolan2005automatically}, 
where the task is to identify whether or not 
a pair of sentences have the same meaning (Figure ~\ref{fig:mrpc}). 
We fine-tune the \texorpdfstring{BERT\textsubscript{BASE}}{} model 
as described in \citet{devlin2018bert}, 
except without weight decay and using SGD with a learning rate of $0.01$ 
instead of the Adam optimizer. 
Our implementation is adapted from \citet{wolf2018pytorch}. 
Experiments were run with importance weights of inverse powers of $4$ for each class, including with no importance weighting. 
All classifiers achieved test accuracies between $78\%$ and $85\%$.

\section{Discussion}
\label{sec:discussion}
Our results show that as training progresses, 
the effects due to importance weighting vanish
(Figures ~\ref{fig:2dsynth},  ~\ref{fig:cifar}, ~\ref{fig:multiclass}, ~\ref{fig:mrpc}
).
While weighting impacts a model's decision boundary early in training,
in the limit, models with widely-varying weights
appear to approach similar solutions. 
After many epochs of training,
there is no clear correlation between the class-based importance weights
and the classification ratios on either test set images, 
out-of-domain images, or random vectors.
When correcting for label shift, IW-ERM confers a benefit early on
that dissipates over epochs of training as the impact of the weighting wears off 
(Figure ~\ref{fig:multiclass}).
In the previous section, we noted that not only do differently-weighted CIFAR models 
converge to similar classification ratios, 
but they also tend to agree on example labels, 
i.e., they learn similar separators.

We show that these findings hold for both simple convolutional networks trained on images,
as well state-of-the-art attention/transformer-based models 
fined-tuned in a transfer learning scheme to text data 
(Figures ~\ref{fig:cifar} and ~\ref{fig:mrpc}). 
These effects are present when training on other pairs of CIFAR-10 classes 
such as cars/trucks (Figure ~\ref{fig:cartruck}), 
and continue to hold when models are optimized by the Adam optimizer, 
although the motivating theory applies to SGD but not ADAM \citep{soudry2017implicit} (Figure ~\ref{fig:adam}).

In contrast, sub-sampling the training set instead of down-weighting the loss function, \textit{does} have a noticeable effect on classification ratios. 
Models assign more CIFAR-10 test images from all classes (both in-domain and out-of-domain)
as well as more random noise images to the majority class.
(Figure ~\ref{fig:subsamp}). 
Notably, weighting the loss function 
to counteract this imbalance during training 
does not balance the classification ratios.

One might note that because the true labeling function is deterministic,
and because our models are sufficiently expressive 
to avoid phenomena due to model misspecification,
perhaps we should not be surprised that import weighting has no effect.
Indeed, for our examples of separable Gaussians and CIFAR images,
we have not altered the Bayes optimal predictions.
To address the matter, we conducted experiments where CIFAR images
are corrupted by label noise. 
Thus the true classes are no longer separable 
but owing to the finite sample and the expressive power of deep nets,
our training data is still nevertheless separable by our model.
Indeed, our experiments showed that under label noise,
IW-ERM still does not meaningfully impact classification ratios (Figure ~\ref{fig:labelnoise}).

In all experiments, models with more extreme weighting 
converge more slowly in decision boundary, 
and convergence in classification ratio begins to occur 
long after perfect training accuracy is achieved. 
For example, the BERT model is typically fine-tuned for $3$ or $4$ epochs \citep{devlin2018bert}, 
however it took over $100$ epochs for the test classification ratios of models 
with more extreme importance weights to stabilise (Figure ~\ref{fig:mrpc}). \citet{lipton2018detecting} trained neural networks with importance weights on CIFAR-10 for $10$ epochs to correct for test label shift, but we find that it can take  up to $200$ epochs for some networks to converge in test accuracy.

An effect of importance weighting on classification ratios 
is present after training ResNet models for $1000$ epochs. 
However, when batch normalization is removed from the model, classification ratios 
during training resemble 
those of the ordinary convolutional network (Figure \ref{fig:resnet}).

We also show that the presence of L2 regularization 
impacts importance-weighted classifiers (Figure \ref{fig:cifar}).    
For the synthetic data, both logistic regression 
and the neural network partition less of the sample space to the down-weighted class (Figure \ref{fig:2d_l2reg}). 
In the CIFAR experiments, L2 regularization slows the convergence in classification ratios of all models (Figure ~\ref{fig:cifar_l2}). 
However, these effects diminish when L2 regularization 
is replaced with dropout (Figures ~\ref{fig:2d_dropout} 
~\ref{fig:cifar}, ~\ref{fig:cifar_do}). In this case, the classifiers behave similar to the unregularized models.

\section{Related Work}
\label{sec:related}
To our knowledge, no previous paper explicitly studies
the effects of importance weighting 
on the decision boundaries learned by modern deep neural networks.
In Section \ref{sec:intro},
we referenced numerous papers applying importance weighting
in a variety of contexts to both to classical and deep models
and in Section \ref{sec:motivation} we referenced those papers 
whose theoretical contributions 
motivated our study.
Here, we briefly recap the most related works.

\paragraph{Theoretical Inspiration}
Our experiments draw inspiration primarily from the works of 
\citet{soudry2017implicit, gunasekar2018implicit},
who proved that deep linear nets are (importance) weight-agnostic
when optimized by SGD to minimize cross entropy loss 
on (linearly) separable data, and the work of \cite{shimodaira2000improving}
which clearly motivates the efficacy of 
importance weighting to model misspecification.

\paragraph{Importance weighting and deep learning}
A number of papers have employed IW-ERM with varying results.
\citet{joachims2018deep} uses deep networks to learn 
from logged contextual bandit feedback,
and \citet{murali2016tsc} weight certain training demonstrations 
in the context of deep imitation learning.
Interestingly, \citet{kostrikov2018addressing}
propose an imitation learning algorithm 
that ought to require importance sampling,
but omit it, noting that empirically the algorithm works regardless.
\citet{schaul2015prioritized} propose a heuristic algorithm 
that up-samples experiences from the replay buffer.
In the case of domain adaptation, 
\citet{azizzadenesheli2019regularized, lipton2018detecting}
use IW-ERM to correct classifiers to account for label shift.
Investigating deep nets for causal inference, \cite{shalit2017estimating} 
weights the loss function to account for the sample size of the treatment group. In curriculum learning \citep{bengio2009curriculum};\citep{matiisen2017teacher};\citep{jiang2015self}, training examples are re-weighted by a teacher during the training process with the objective of improving or accelerating training.

\section{Conclusions}
\label{sec:conclusions}
Our experiments suggest that effects from importance weighting on deep networks 
may only occur in conjunction with early stopping, disappearing asymptotically. 
For these over-parameterized models, capable of fitting any training set,
the learned solution may be determined solely by the location of training examples, 
independent of their density. 
For example, when correcting for label shift in test data, 
test accuracy may deteriorate over training epochs even as the classifier improves
owing to the diminishing effect of importance weighting.
Not only do we fail to find any clear correlation between importance weighting 
and the fraction of test examples partitioned to each class, 
but models with different importance weightings also have 
high agreement even on out-of-domain images, 
providing further evidence that the learned decision boundaries are similar.
Our findings should raise concerns amongst practitioners
who might re-evaluate its use on the various problems 
for which importance weighting is a standard tool.

We find similar patterns across various models 
(MLPs, convolutional networks, and attention-based transformer networks)
and domains (synthetic 2D data, images, and natural language).
While importance weighting does appear to have some effect 
when applied with residual networks
we observe that these effects vanish when batch normalization is removed. 
Batch normalization counteracts the effect of exploding weight norms
by normalizing the magnitude of weights for all but the final classification layer.
However, in our experiments, we observe that models with batch normalization, 
still have large final-layer weights, resulting in large logit values after training. 
Thus we speculate that it may be possible for batch normalization 
to interact with importance weighting by some other mechanism.

Some effect of importance weighting can be realized 
when applied in combination with L2 regularization. 
We believe that in this case, the L2 penalty prevents SGD 
from reaching the large norm solutions 
whose loss is dominated by the support vectors, 
thus preventing convergence to max-margin-like solutions. 
This aligns with our related finding that dropout,
which does not penalize such large-norm solutions,
does not affect the fractions of examples partitioned to each class 
(for importance-weighted classifiers) in the limit.

We find that weighting the loss function of deep networks 
fails to correct for training set class imbalance. 
However, sub-sampling a class in the training set 
clearly affects the network's predictions. 
This finding indicates that perhaps sub-sampling 
can be an alternative to importance weighting 
for deep networks on sufficiently large training sets.

While as previously noted, importance weighting has been shown (empirically)
to be useful for deep networks by several others 
\citep{lipton2018detecting, schaul2015prioritized, burda2015importance},
our findings nevertheless support rethinking 
the standard application of importance weighting in combination with deep learning, 
suggesting that practitioners should exercise caution 
when making use of them and raising new questions such as:
if importance weighting is only useful for deep networks 
in conjunction with early stopping or weight decay, 
then is there a principled way to choose stopping times 
or weight decay coefficients when importance weighting is desired?

\section*{Acknowledgments}
We would like to thank Daniel Soudry, Ferenc Huszár, Ben Eysenbach,
and the reviewers for providing valuable feedback
that has helped to improve the draft.
We would also like to thank Amazon, Salesforce, Adobe, 
the AI Ethics and Governance Fund,
and the Center for Machine Learning in Health,
whose generous support has helped to advance this line of research
on robust machine learning.

\bibliographystyle{icml2019}
\bibliography{refs}

\clearpage
\appendix
\counterwithin{figure}{section}
\section{Supplemental Materials}
\label{appendix}
\noindent\begin{minipage}{\textwidth}
    \centering
    \captionsetup{type=figure}
    \includegraphics[width=0.95\linewidth]{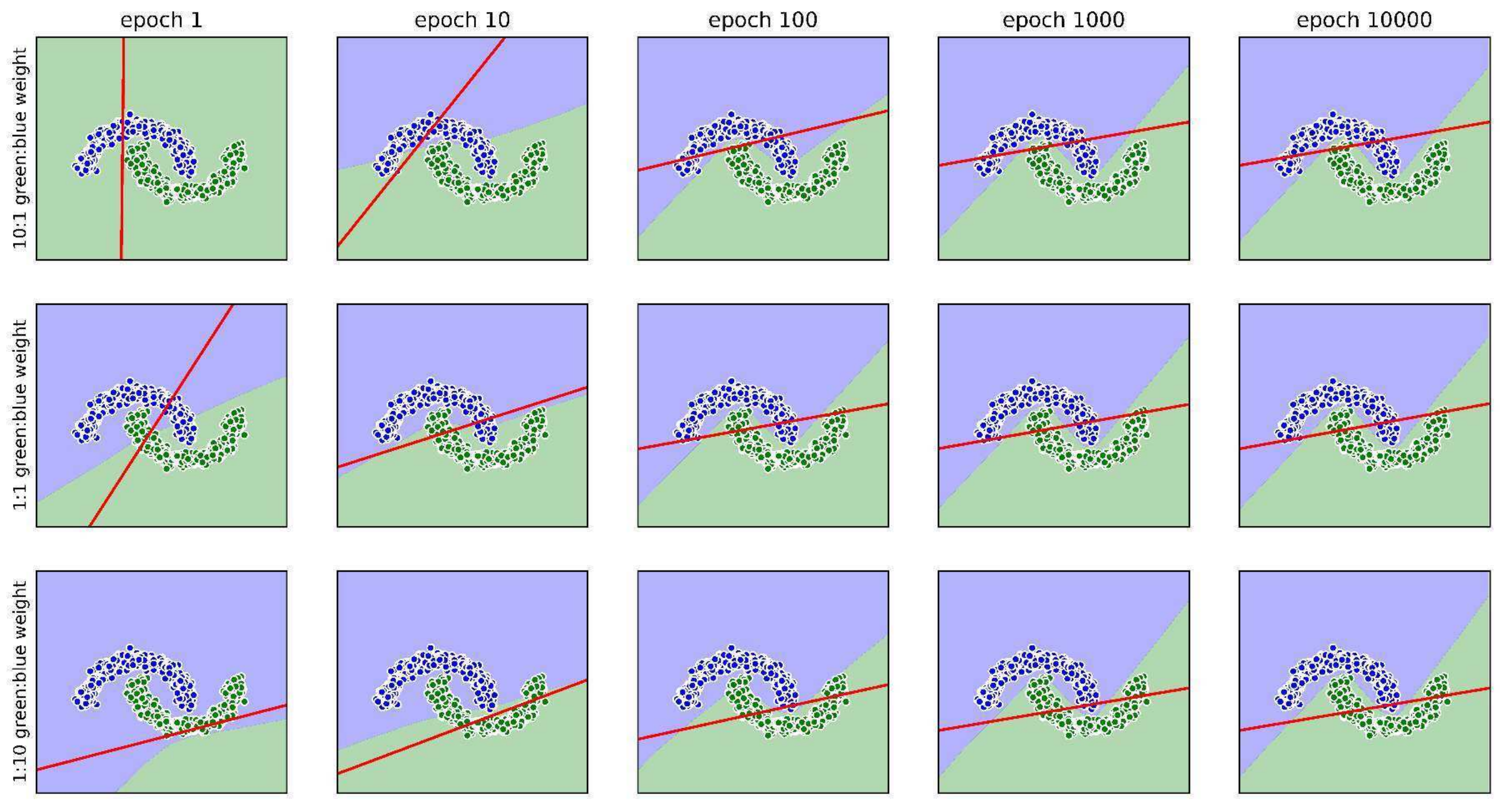}
    \captionof{figure}{Results on non linearly-separable moons dataset.
    \label{fig:2d_moons}}
    
    \vskip 0.3in
    \centering
    \includegraphics[width=0.95\linewidth]{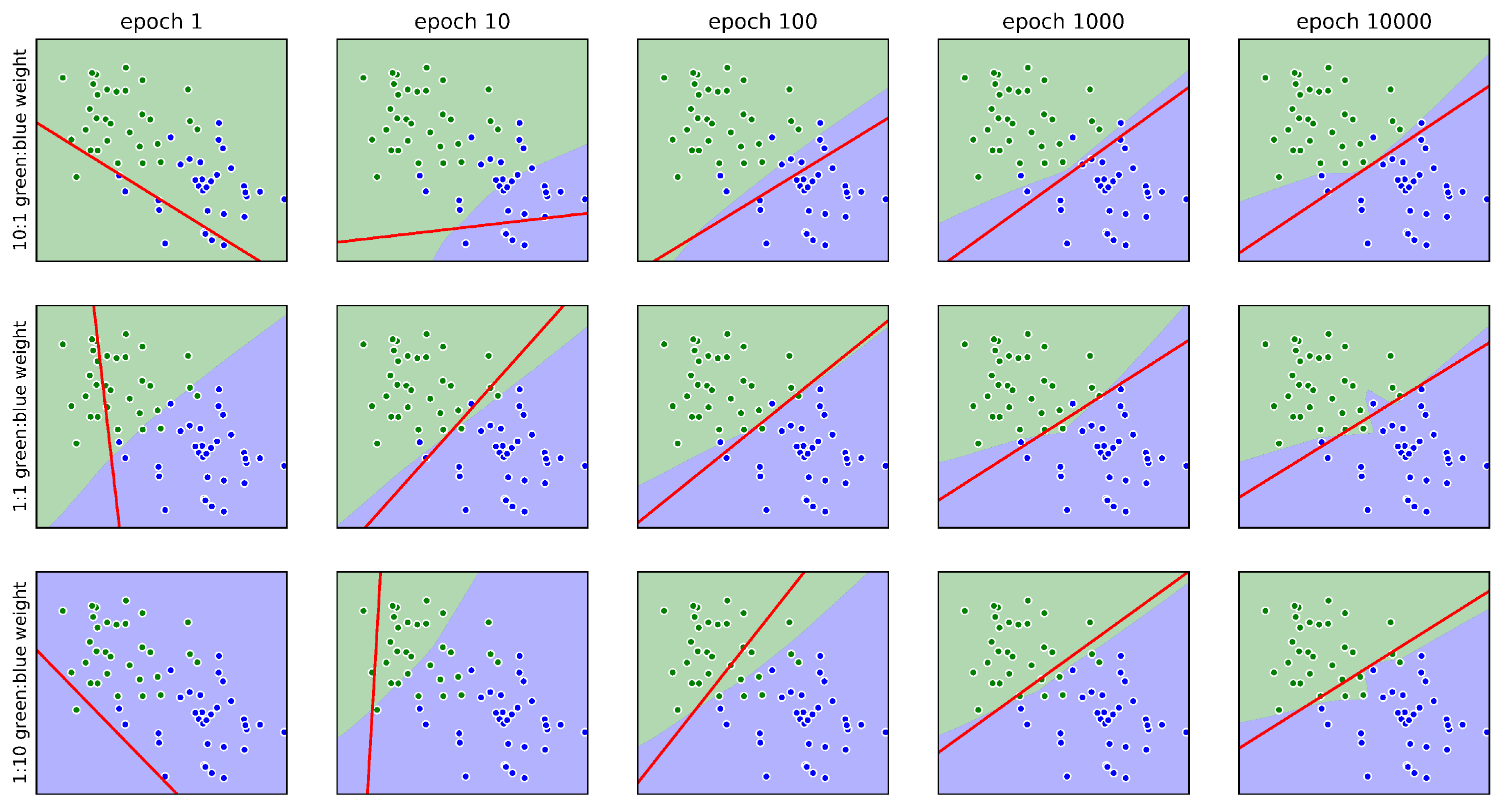}
    \captionof{figure}{Results on overlapping Gaussians dataset.
    \label{fig:2d_noise}}
\end{minipage}



\begin{figure*}[!t]
    \centering
    \subcaptionbox{CIFAR10 cat and dog test images.}%
    [.32\textwidth]{\includegraphics[width=.32\textwidth]{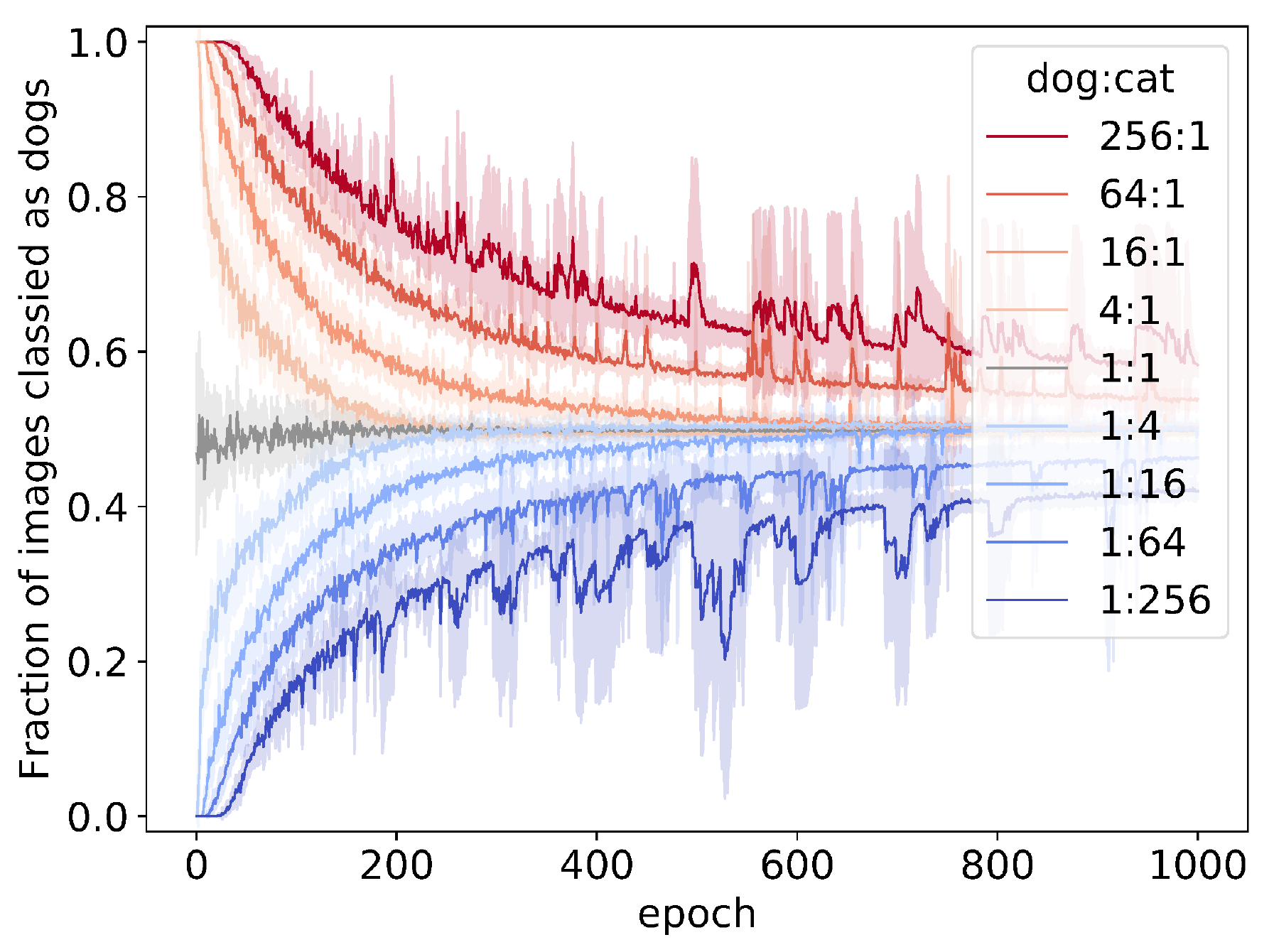}}
    \label{fig:cifar_test_l2}
    \hfill
    \subcaptionbox{CIFAR10 test images from the other eight classes.}%
    [.32\textwidth]{\includegraphics[width=.32\textwidth]{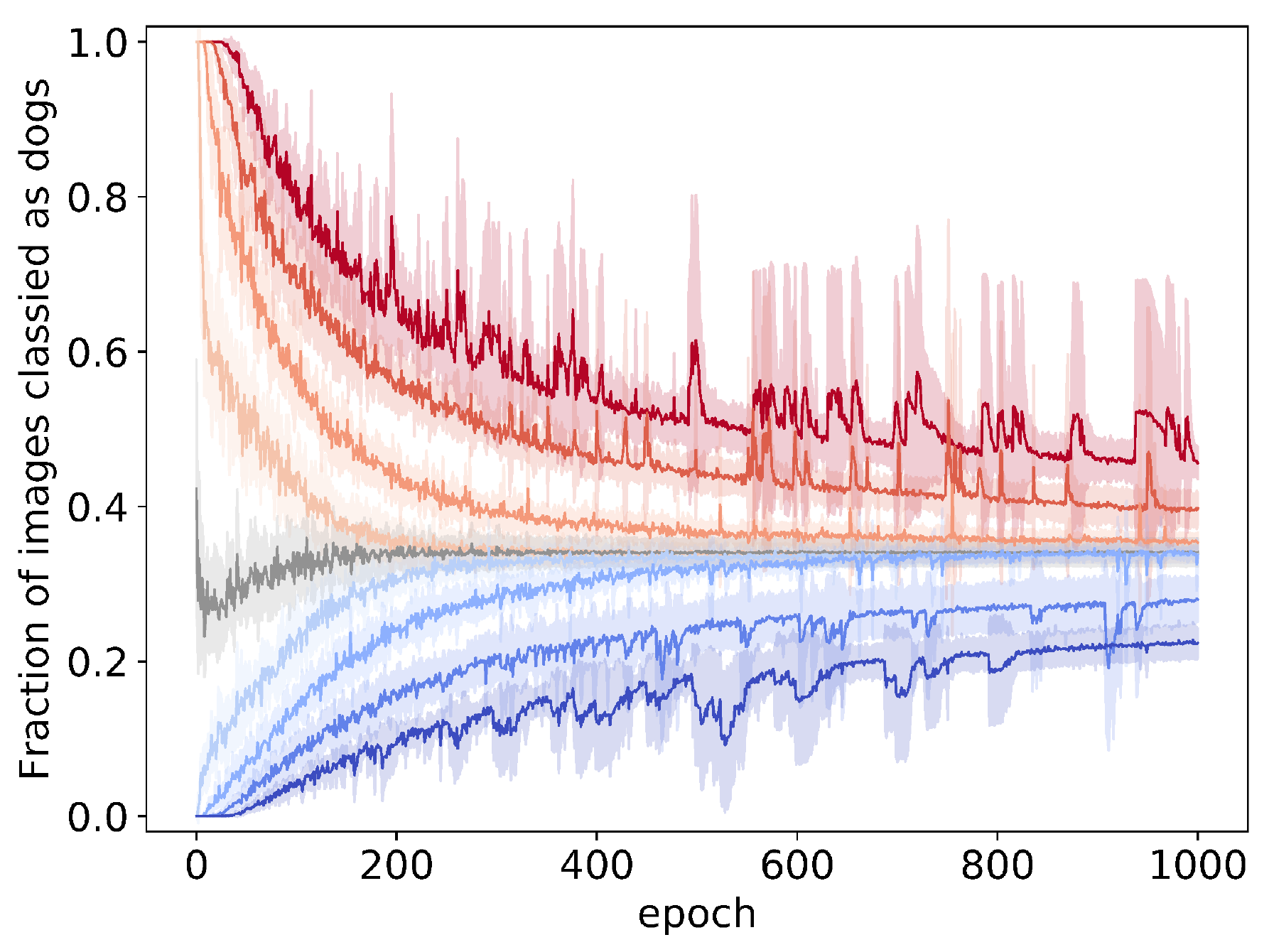}}
    \label{fig:cifar_other_l2}
    \subcaptionbox{Random images.}%
  [.32\textwidth]{\includegraphics[width=.32\textwidth]{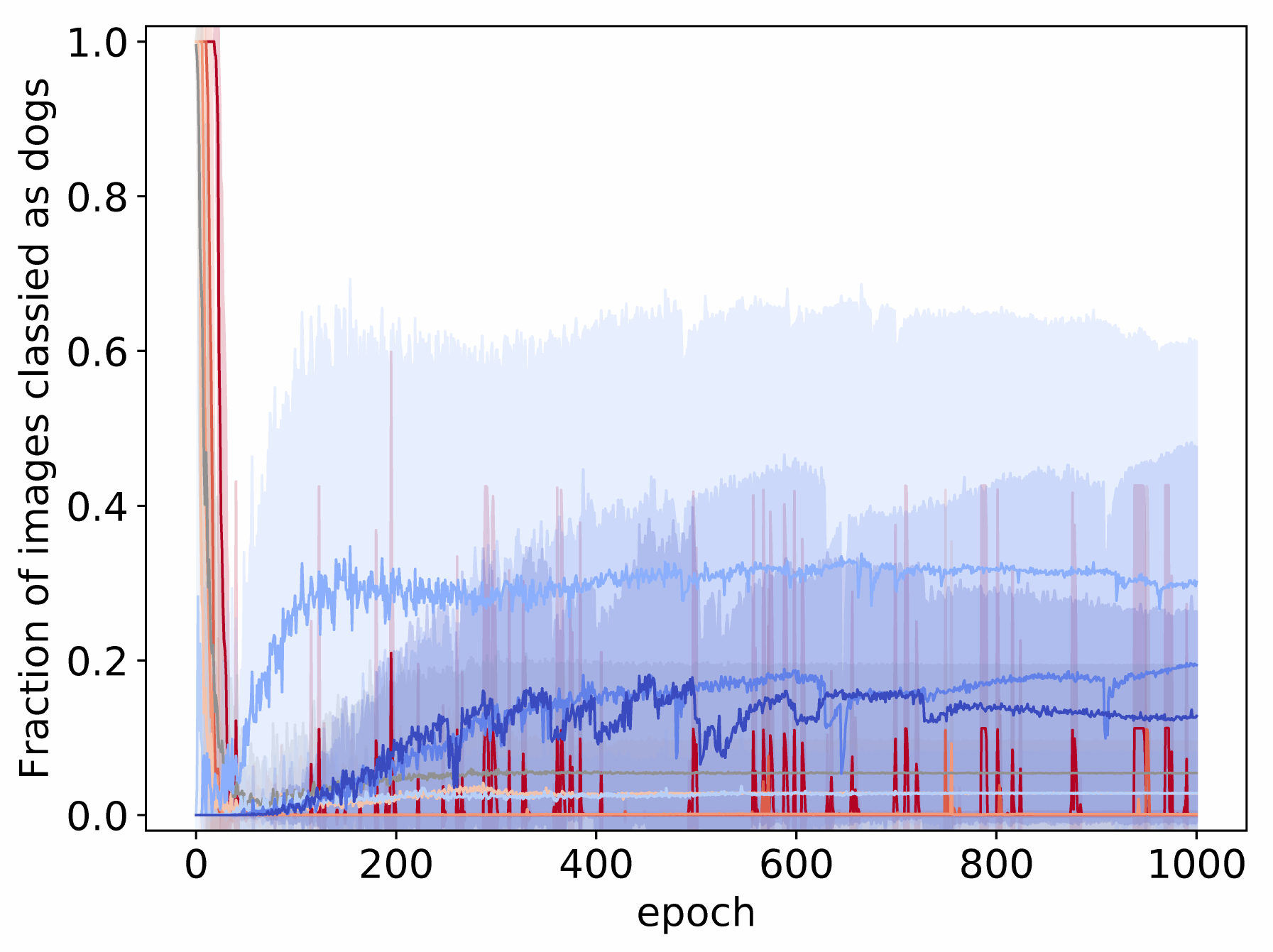}}
    \label{fig:cifar_random_l2}
    \caption
    {Effect of early stopping on L2 regularized models. Plots are structured as in Figure ~\ref{fig:cifar}.} 
    \label{fig:cifar_l2}
\end{figure*}

\begin{figure*}[!t]
    \centering
    \subcaptionbox{CIFAR10 cat and dog test images.}%
    [.32\textwidth]{\includegraphics[width=.32\textwidth]{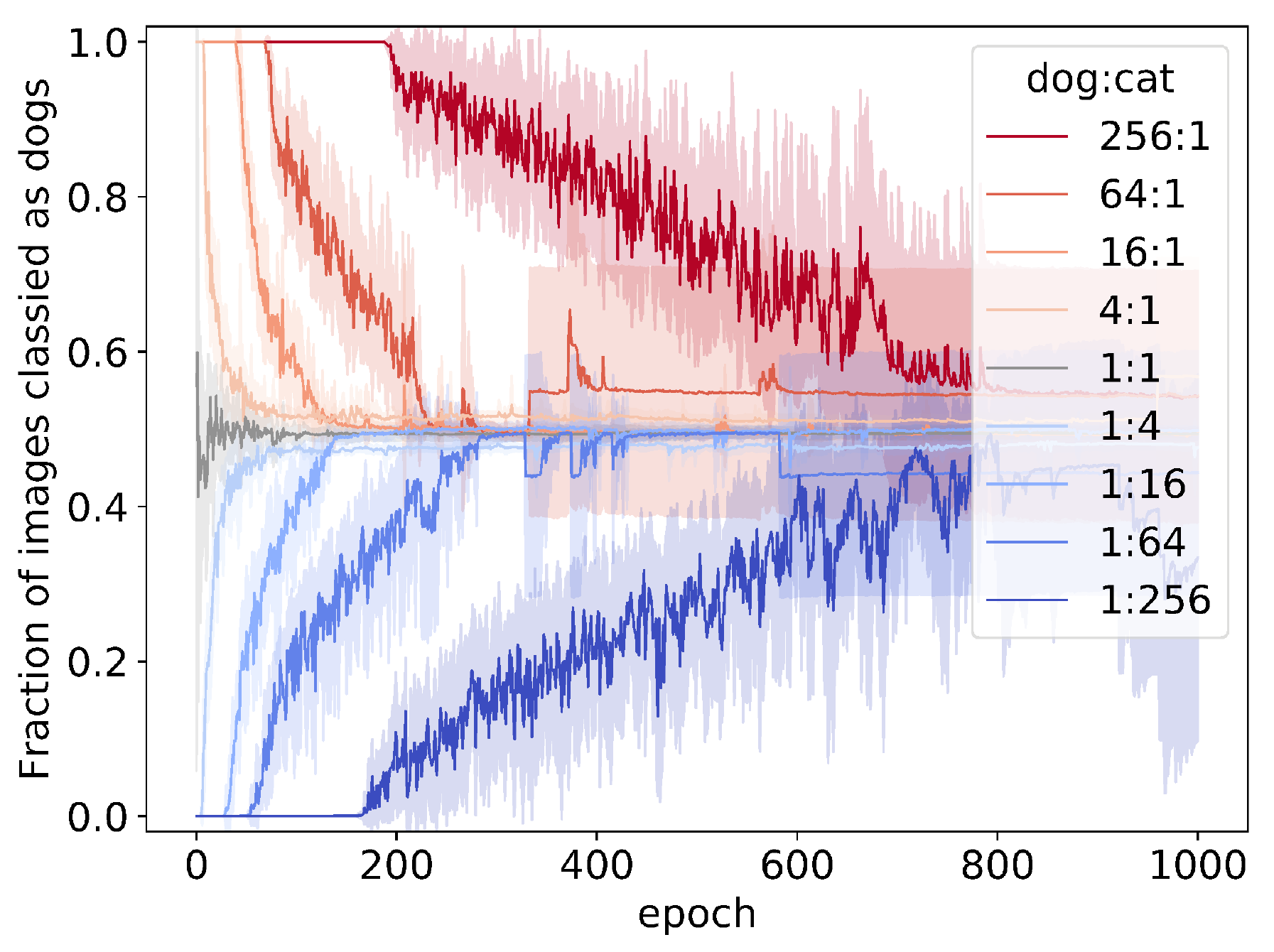}}
    \label{fig:cifar_test_dropout}
    \hfill
    \subcaptionbox{CIFAR10 test images from the other eight classes.}%
    [.32\textwidth]{\includegraphics[width=.32\textwidth]{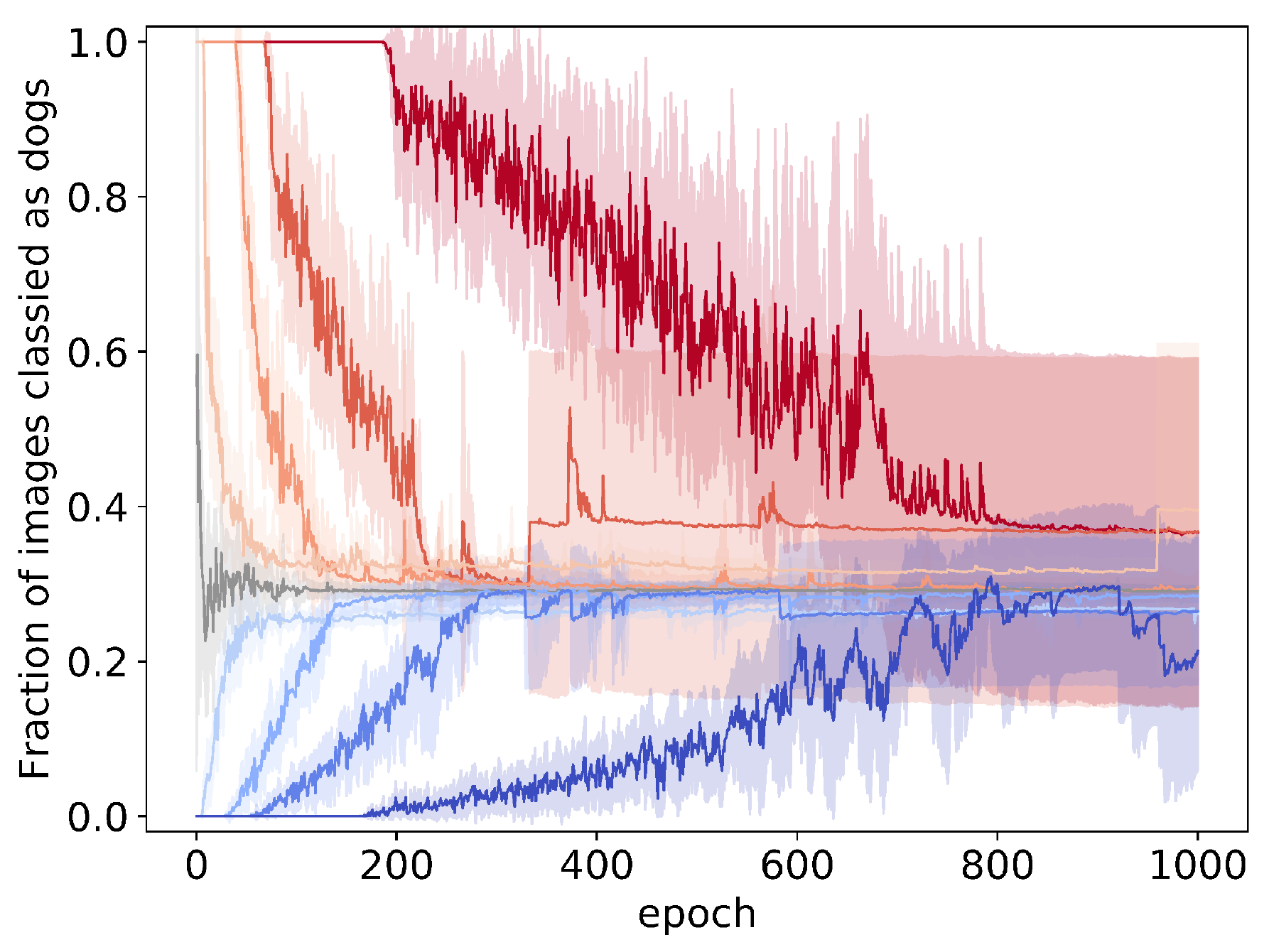}}
    \label{fig:cifar_other_dropout}
    \subcaptionbox{Random images.}%
  [.32\textwidth]{\includegraphics[width=.32\textwidth]{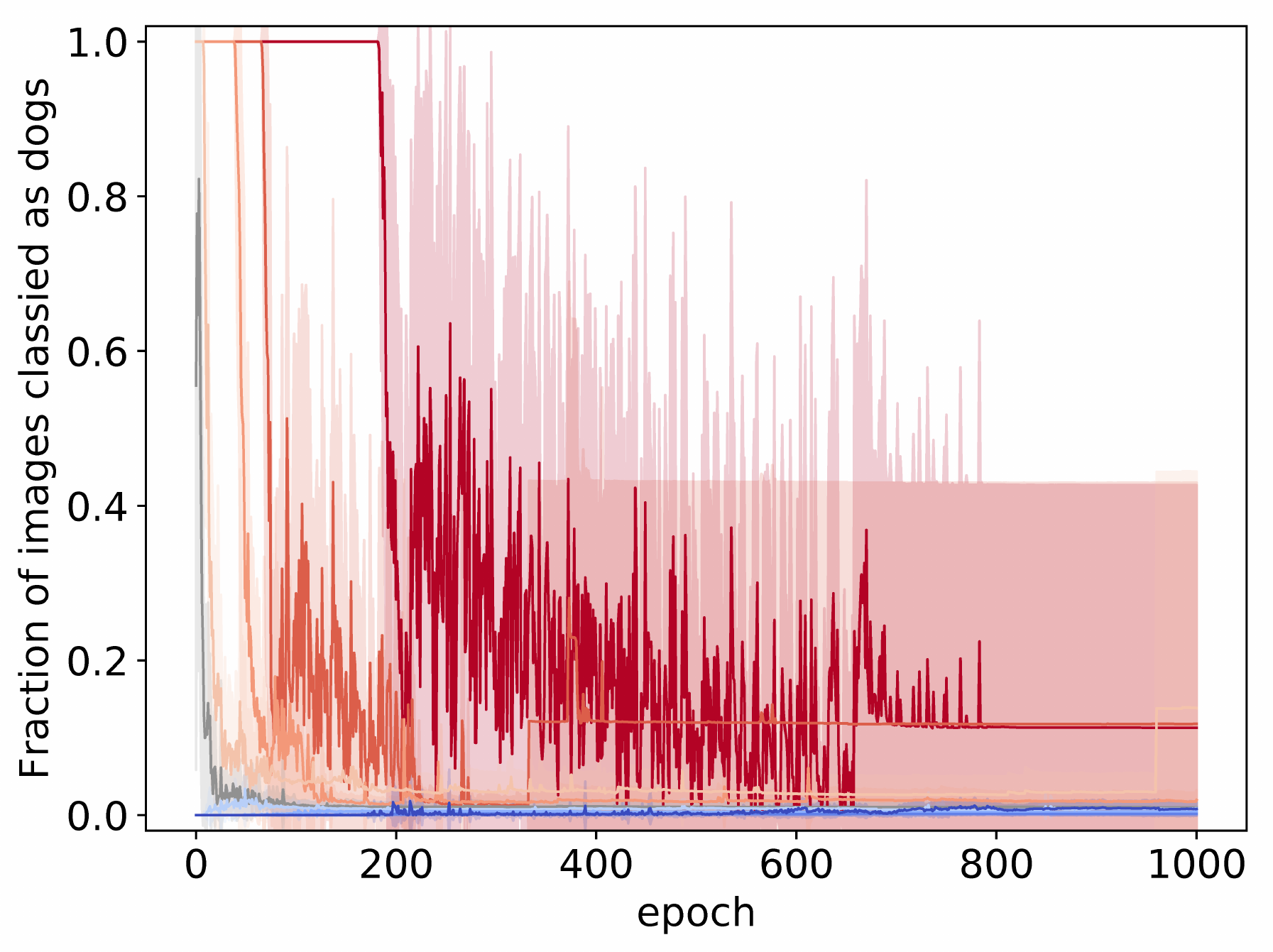}}
    \label{fig:cifar_random_dropout}
    \caption
    {Effect of early stopping on models with dropout. 
    } 
    \label{fig:cifar_do}
\end{figure*}

\begin{figure*}[!t]
    \centering
    \subcaptionbox{ResNet with batchnorm.}%
    [.47\textwidth]{\includegraphics[width=.45\textwidth]{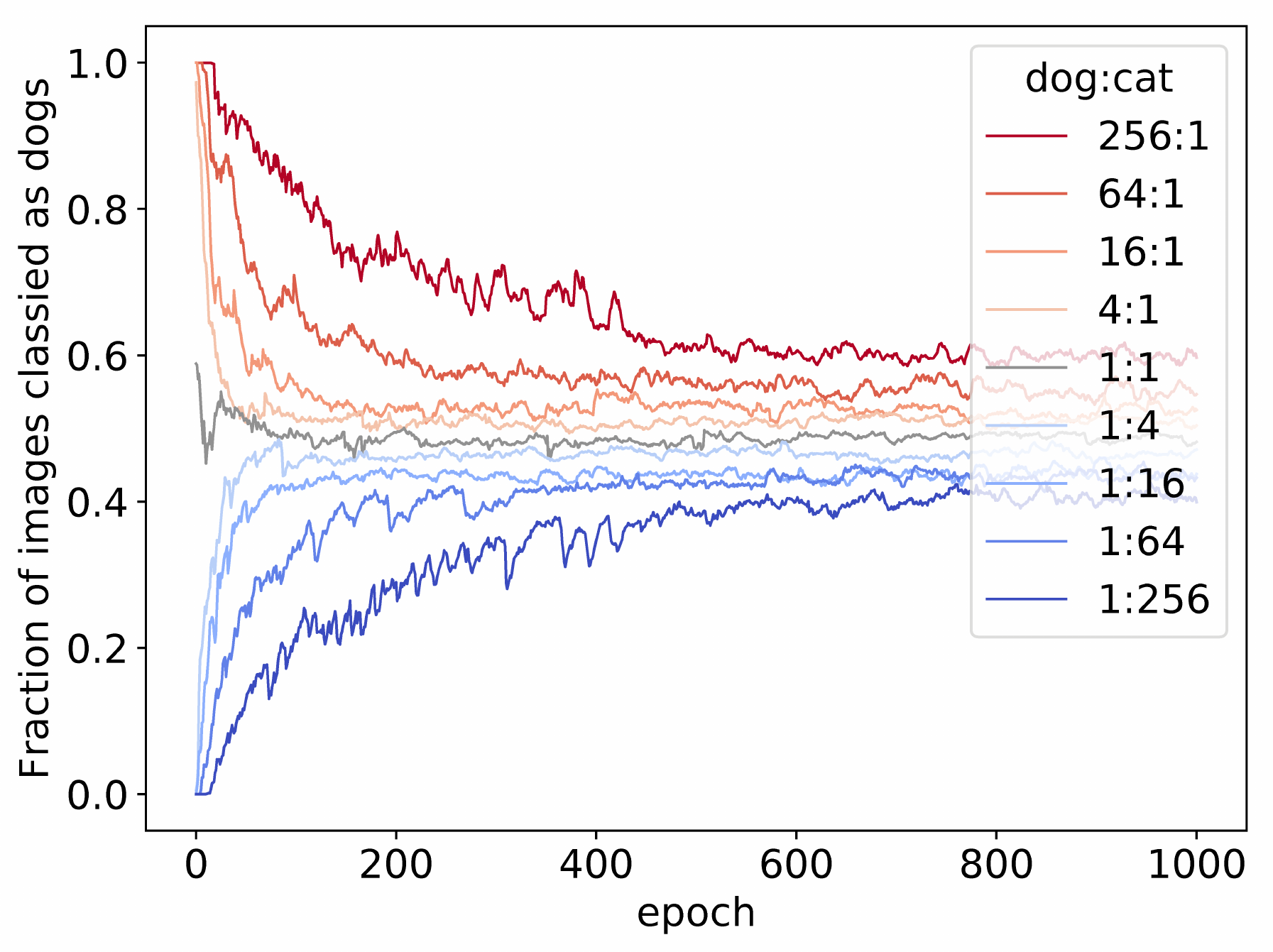}}
    \vspace{-5px}
    \label{fig:resnetepoch}
    \hfill
    \subcaptionbox{ResNet without batchnorm.}%
    [.47\textwidth]{\includegraphics[width=.45\textwidth]{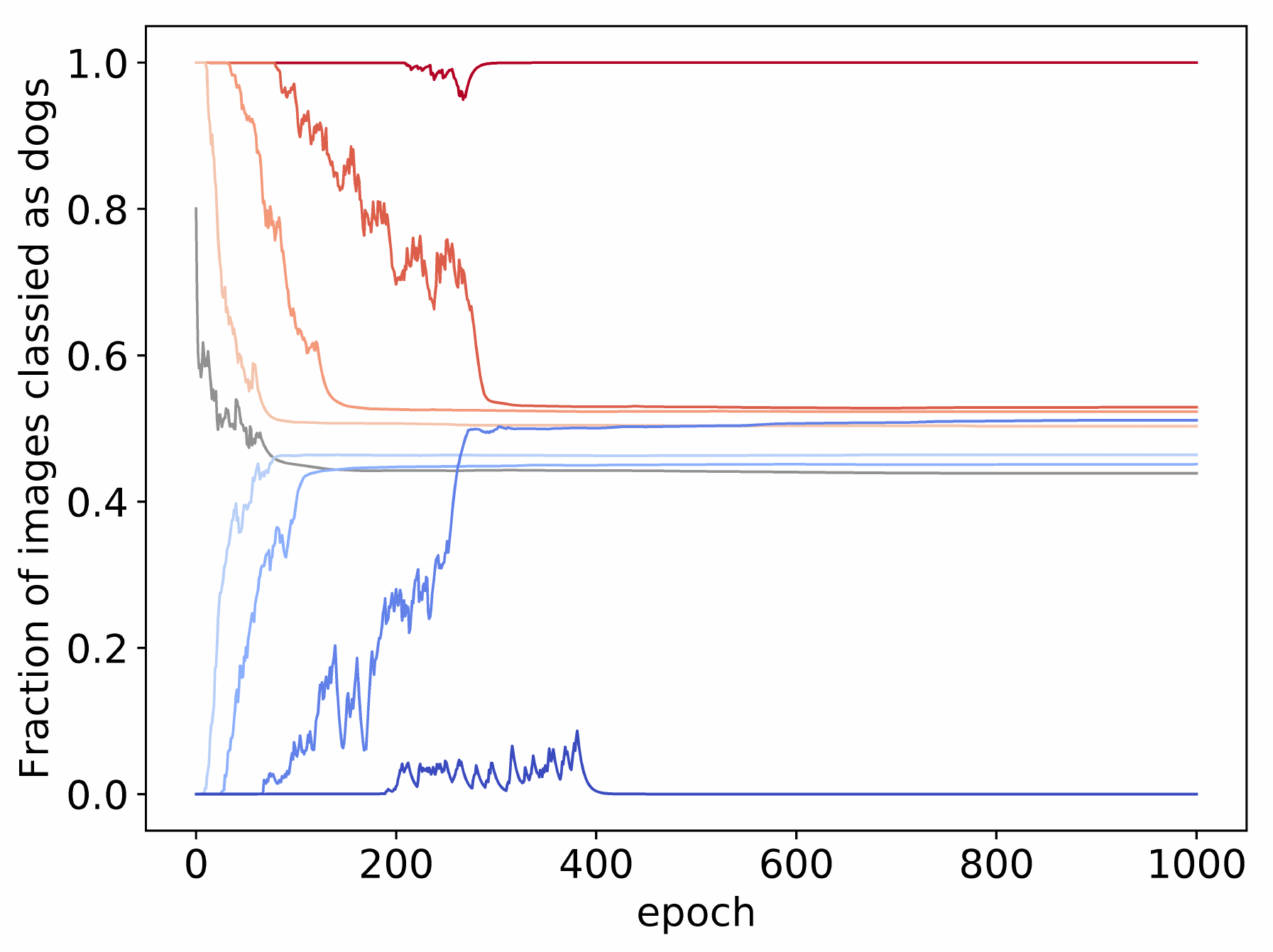}}
    \vspace{-5px}
    \label{fig:resnetepochnobn}
    \vskip\baselineskip
    \subcaptionbox{ResNet with batchnorm.}%
    [.47\textwidth]{\includegraphics[width=.45\textwidth]{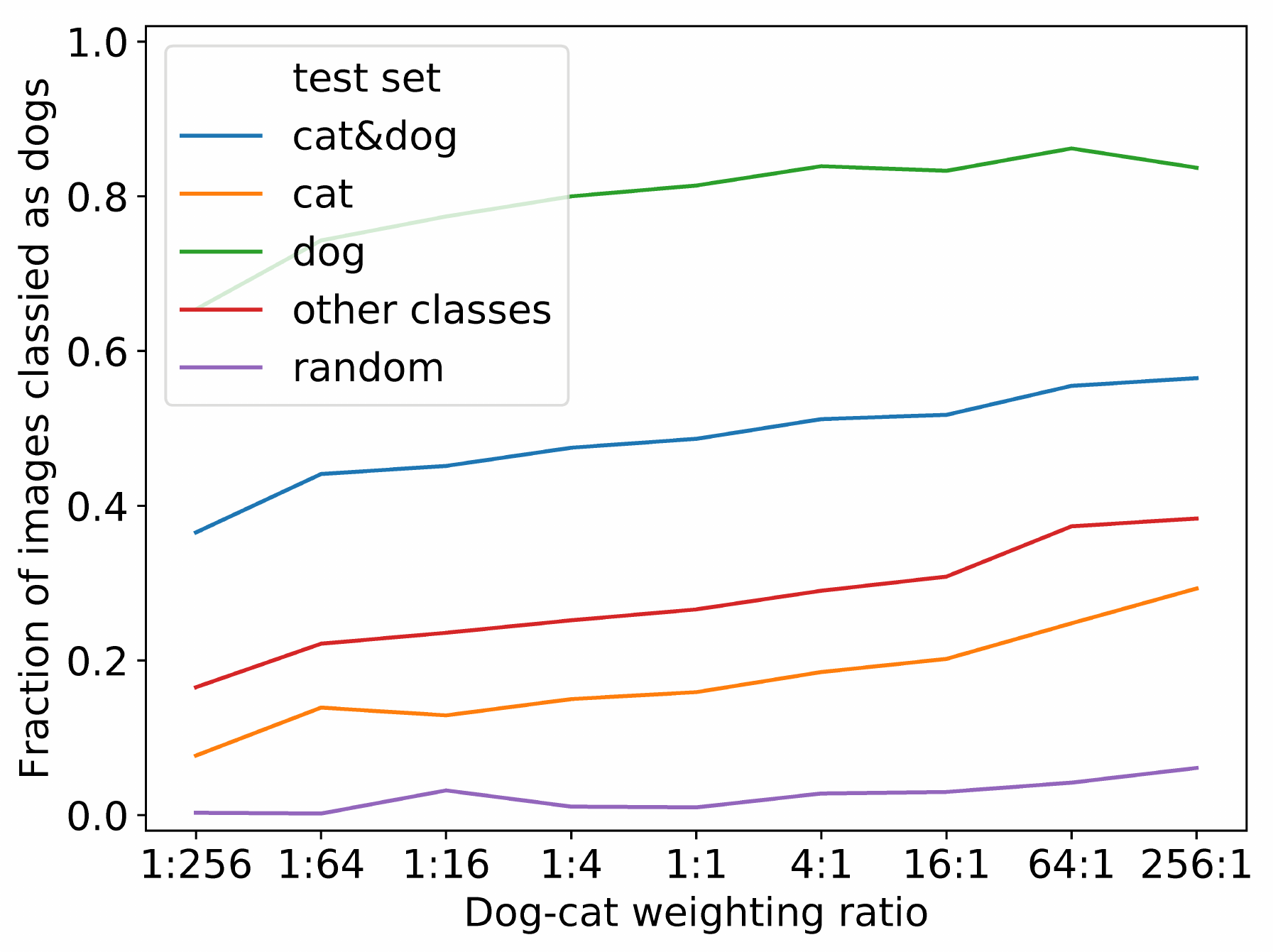}}
    \label{fig:resnetiw}
    \hfill
    \subcaptionbox{ResNet without batchnorm.}
    [.47\textwidth]{\includegraphics[width=.45\textwidth]{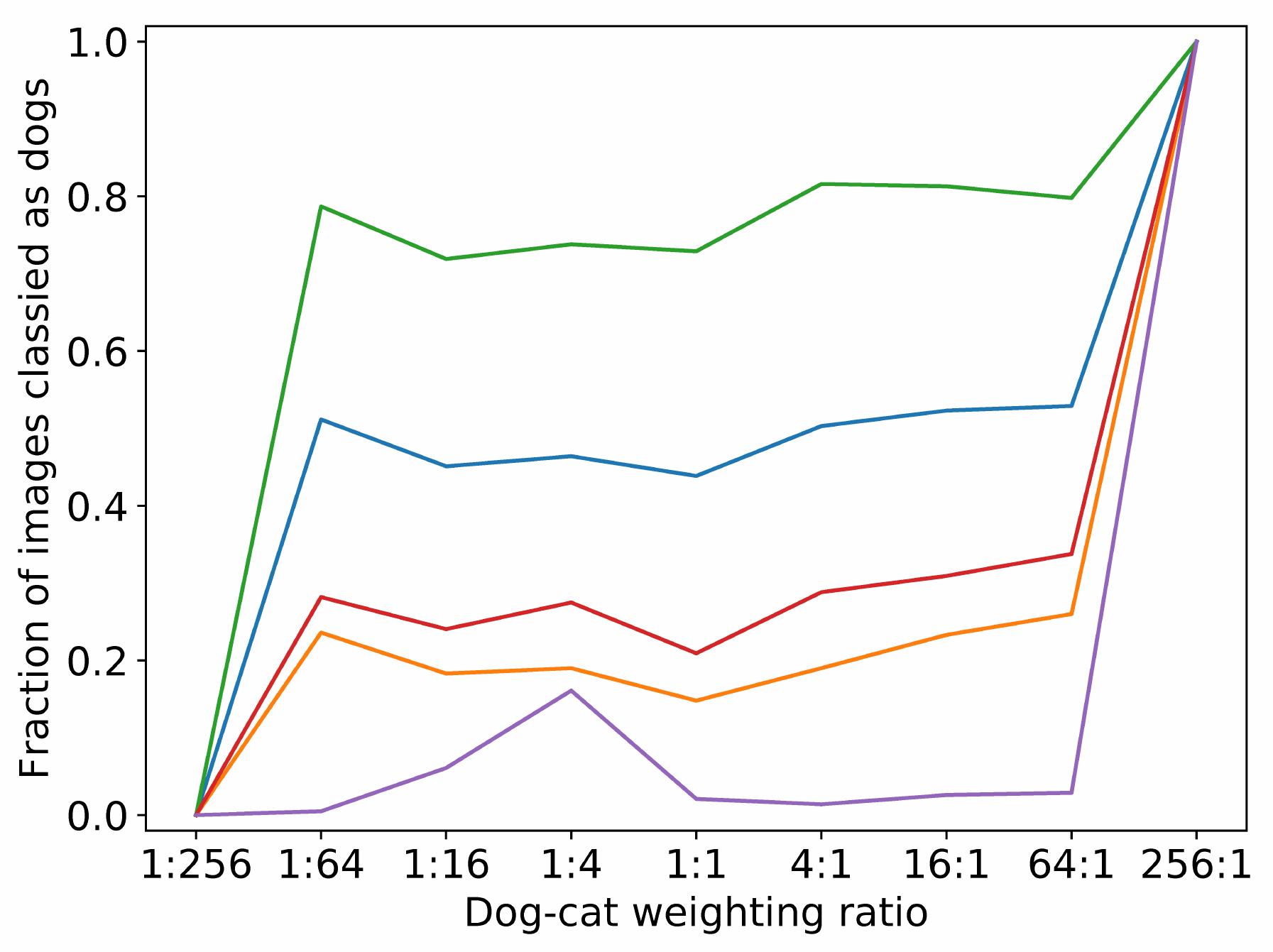}}
    \label{fig:resnetiwnobn}
    \caption
    {Relationship between early stopping and importance weighting (a,b), and final classification ratios vs. importance weighting (c,d) for ResNet models on CIFAR with and without batch normalization. Plots are structured as in Figure ~\ref{fig:cifar}.} 
    \label{fig:resnet}
\end{figure*}

\begin{figure*}[!t]
    \centering
    \subcaptionbox{Without rebalancing via importance weighting.}%
    [.47\textwidth]{\includegraphics[width=.47\textwidth]{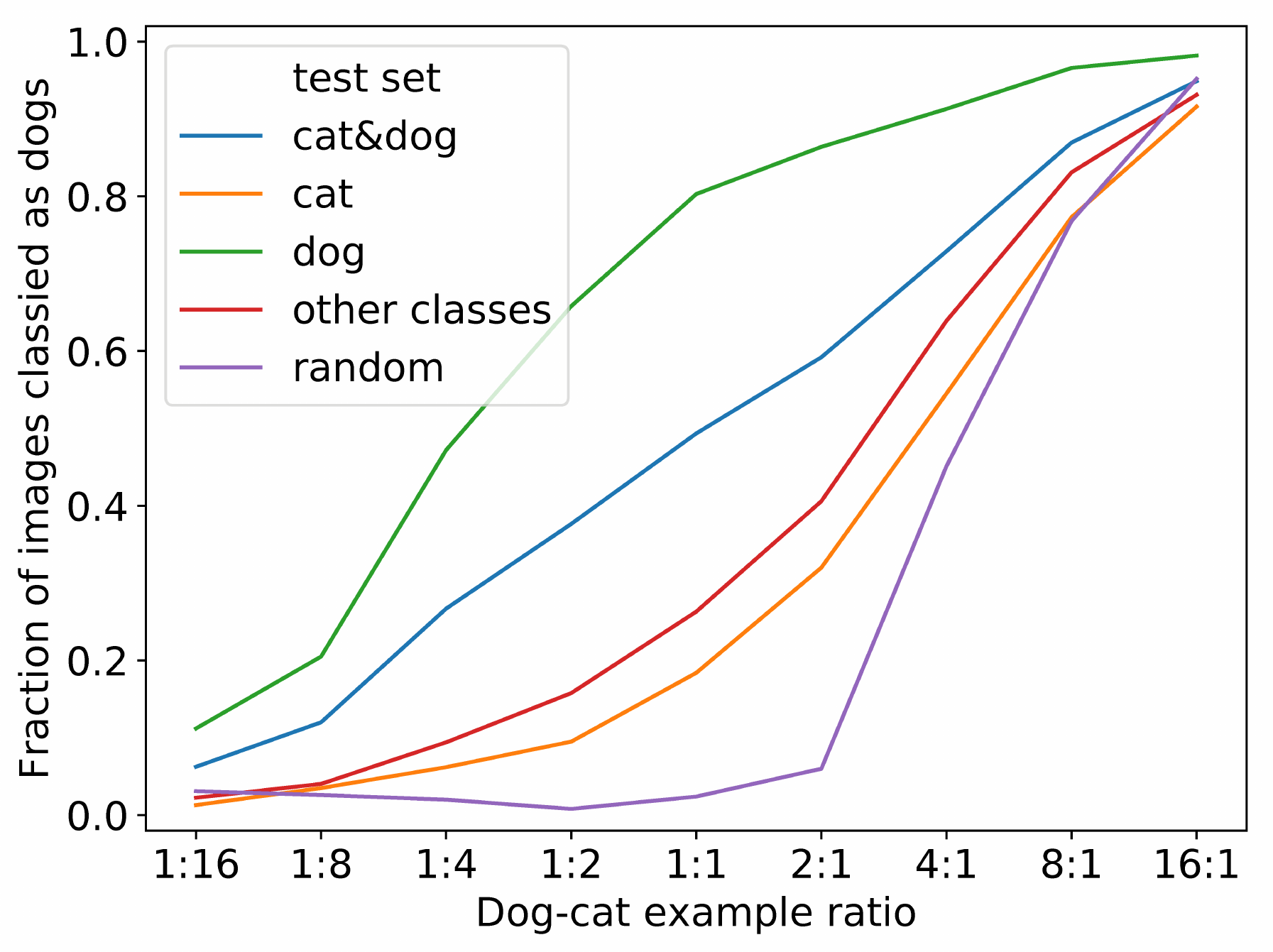}}
    \label{fig:cifar_subsamp_noreweight}
    \hfill
    \subcaptionbox{With rebalancing via importance weighting.}%
    [.47\textwidth]{\includegraphics[width=.47\textwidth]{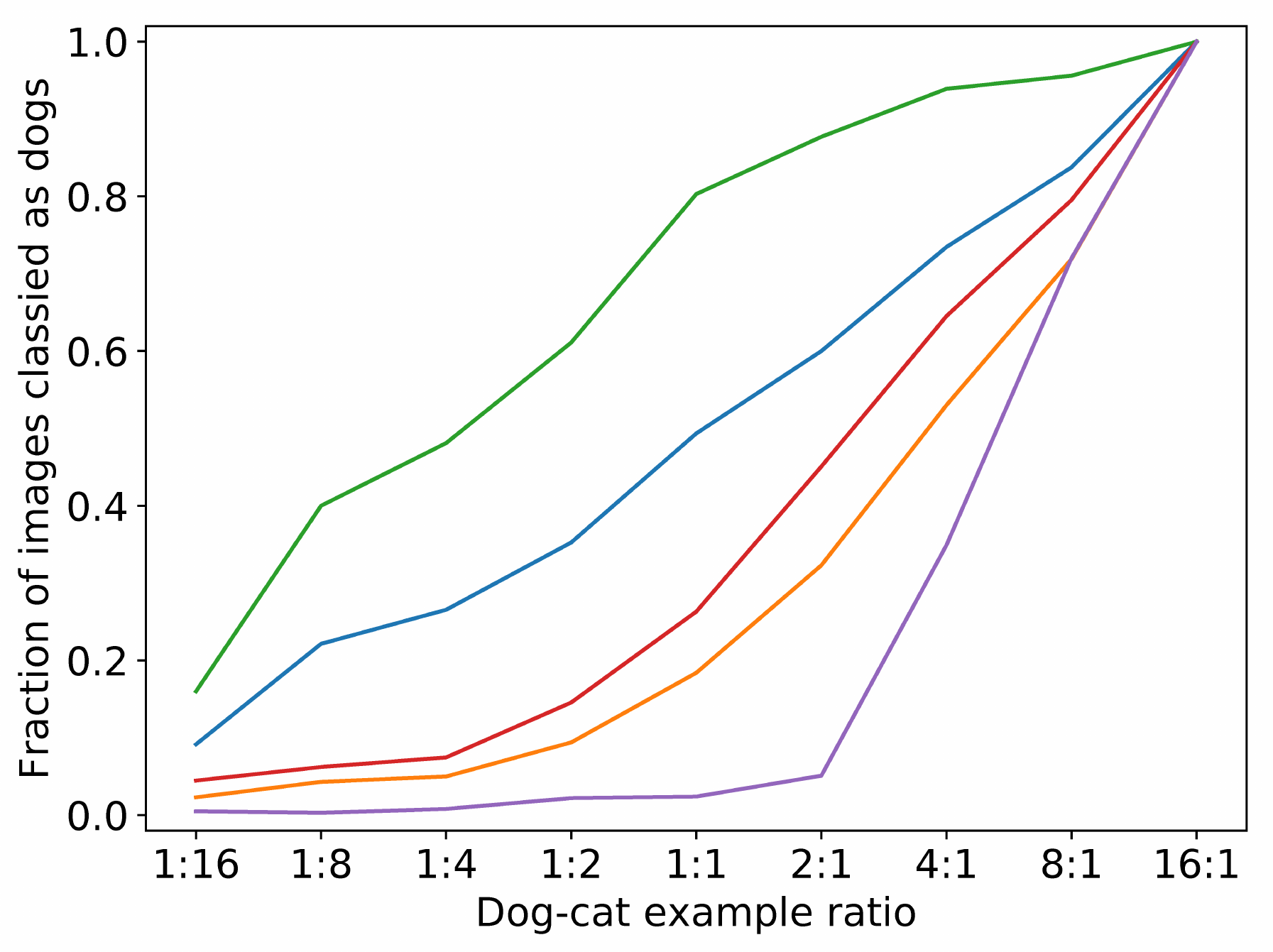}}
    \label{fig:cifar_subsamp_reweight}
    \caption
    {Attempting to correct for class imbalance with importance weights. Results are shown for different sub-sampling ratios (X-axis). Our results suggest that while class balance at training time impacts the classification ratios, IW-ERM does not.}
    \label{fig:subsamp}
\end{figure*}

\begin{figure*}[!t]
    \centering
    \includegraphics[width=.47\textwidth]{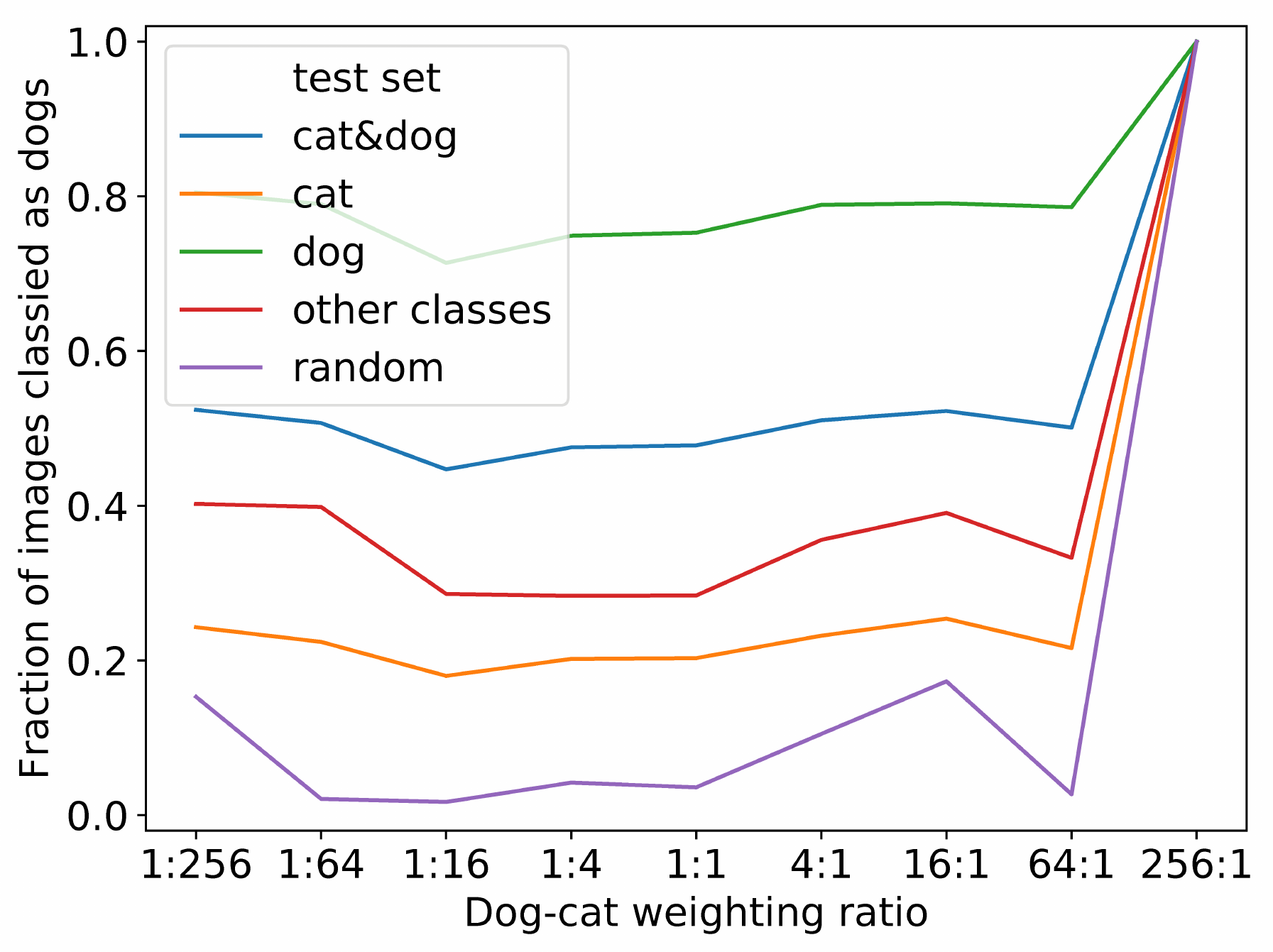}
    \caption
    {Classification ratios vs importance weights on CIFAR10 where the labels of $5\%$ of each class in the training set are flipped. This experiments shows that even when the true classes are inseparable, importance weighting has surprisingly little effect so long as the neural network can separate the training data.} 
    \label{fig:labelnoise}
\end{figure*}

\begin{figure*}[!t]
    \centering
    \subcaptionbox{CIFAR10 car and truck test images.}%
    [.47\textwidth]{\includegraphics[width=.47\textwidth]{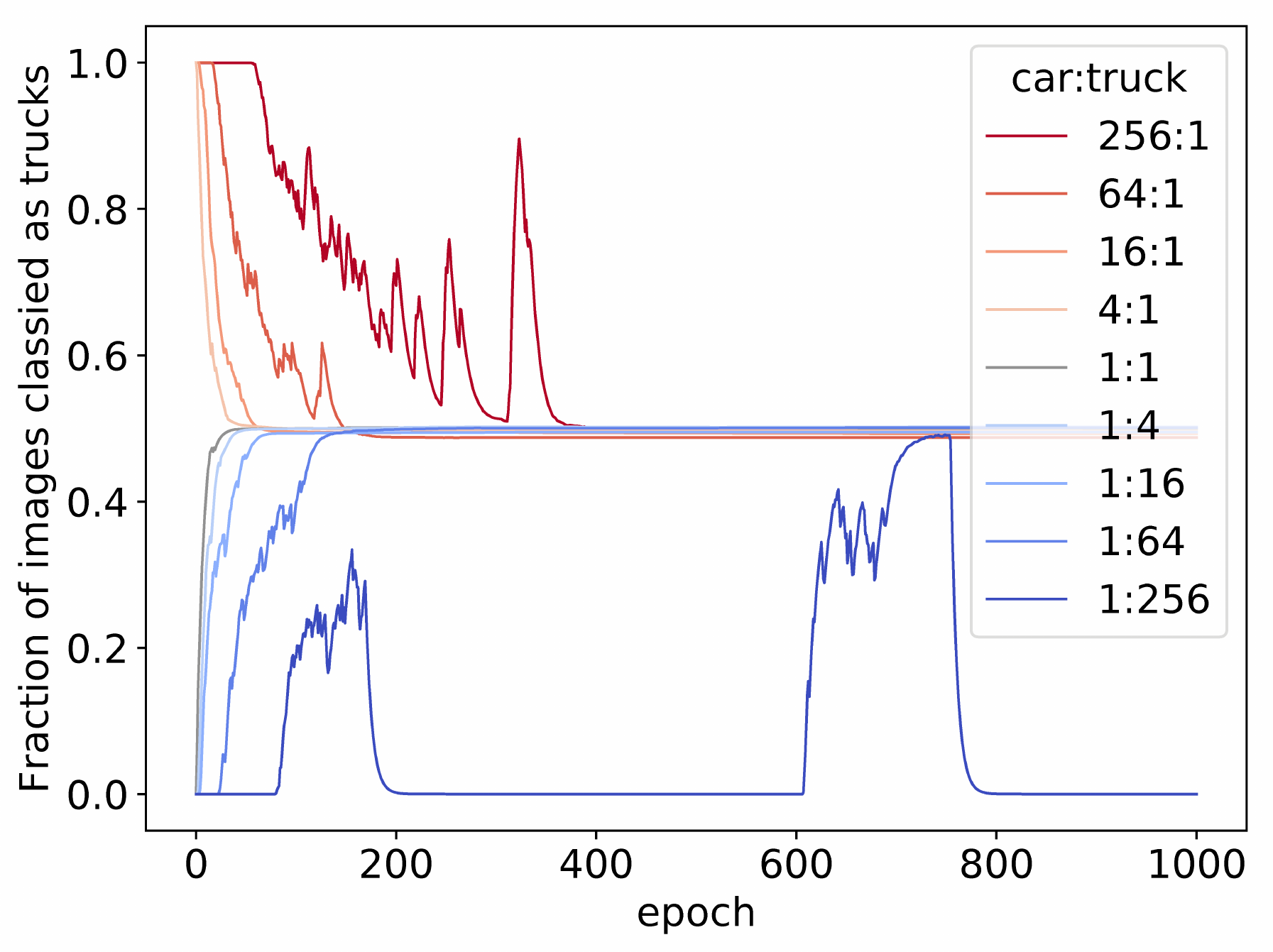}}
    \label{fig:cifar_cartruck_test}
    \hfill
    \subcaptionbox{CIFAR10 test images from non-car/truck classes.}%
    [.47\textwidth]{\includegraphics[width=.47\textwidth]{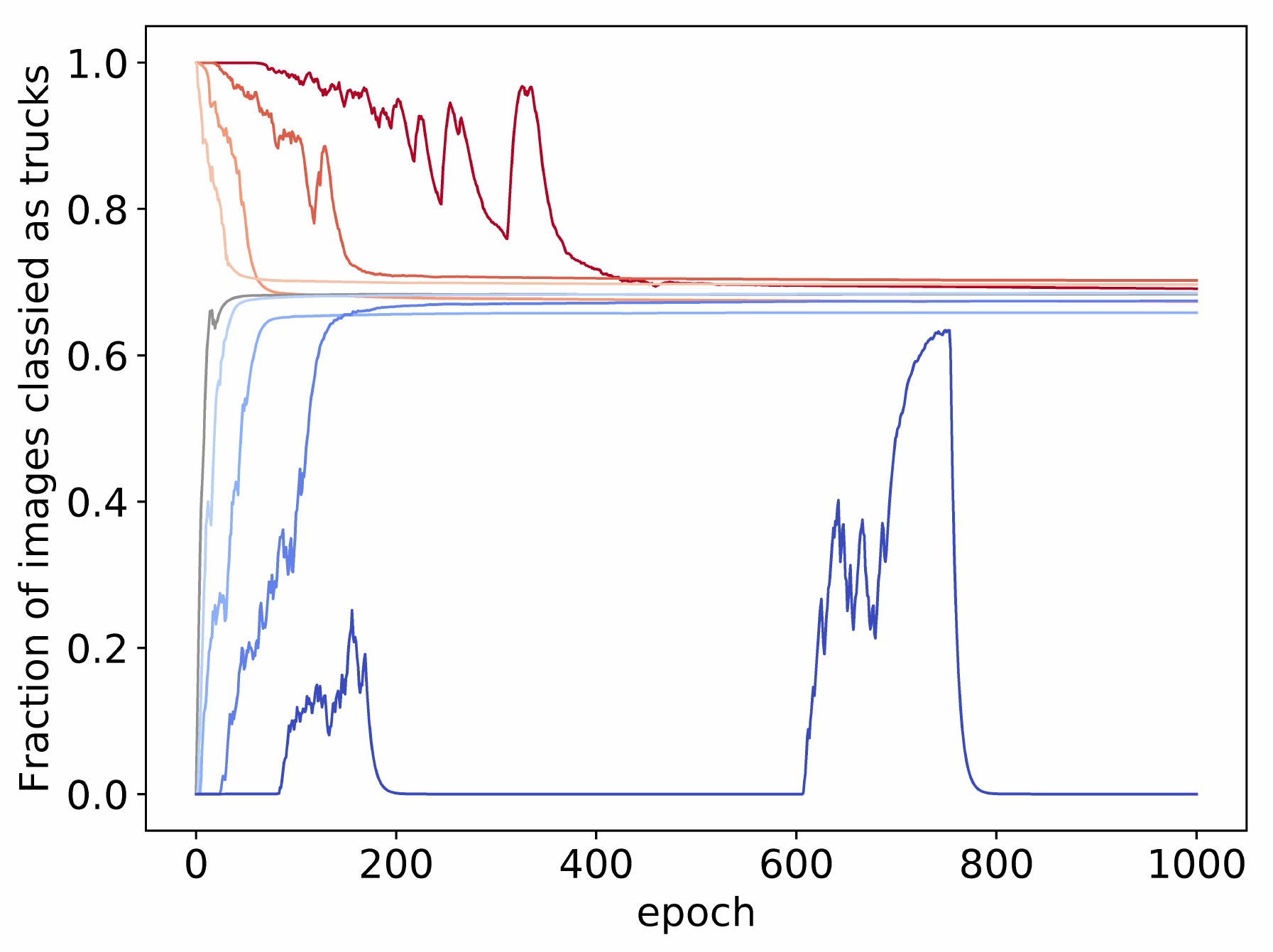}}
    \label{fig:cifar_cartruck_other}
    \vskip\baselineskip
    \subcaptionbox{Random images.}%
    [.47\textwidth]{\includegraphics[width=.47\textwidth]{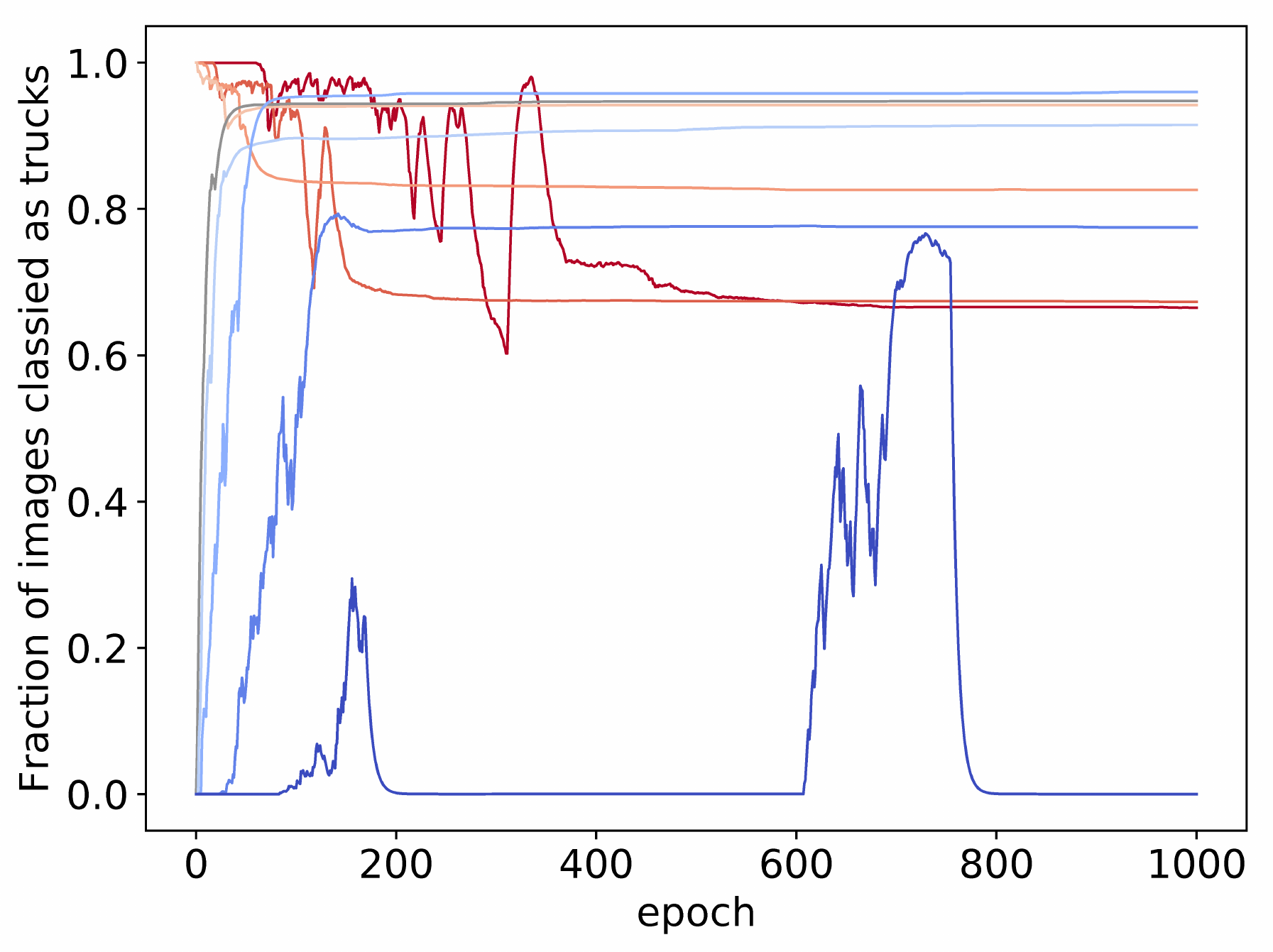}}
    \label{fig:cifar_cartruck_random}
    \hfill
    \subcaptionbox{Classification ratios after training on car/truck images.}%
    [.47\textwidth]{\includegraphics[width=.47\textwidth]{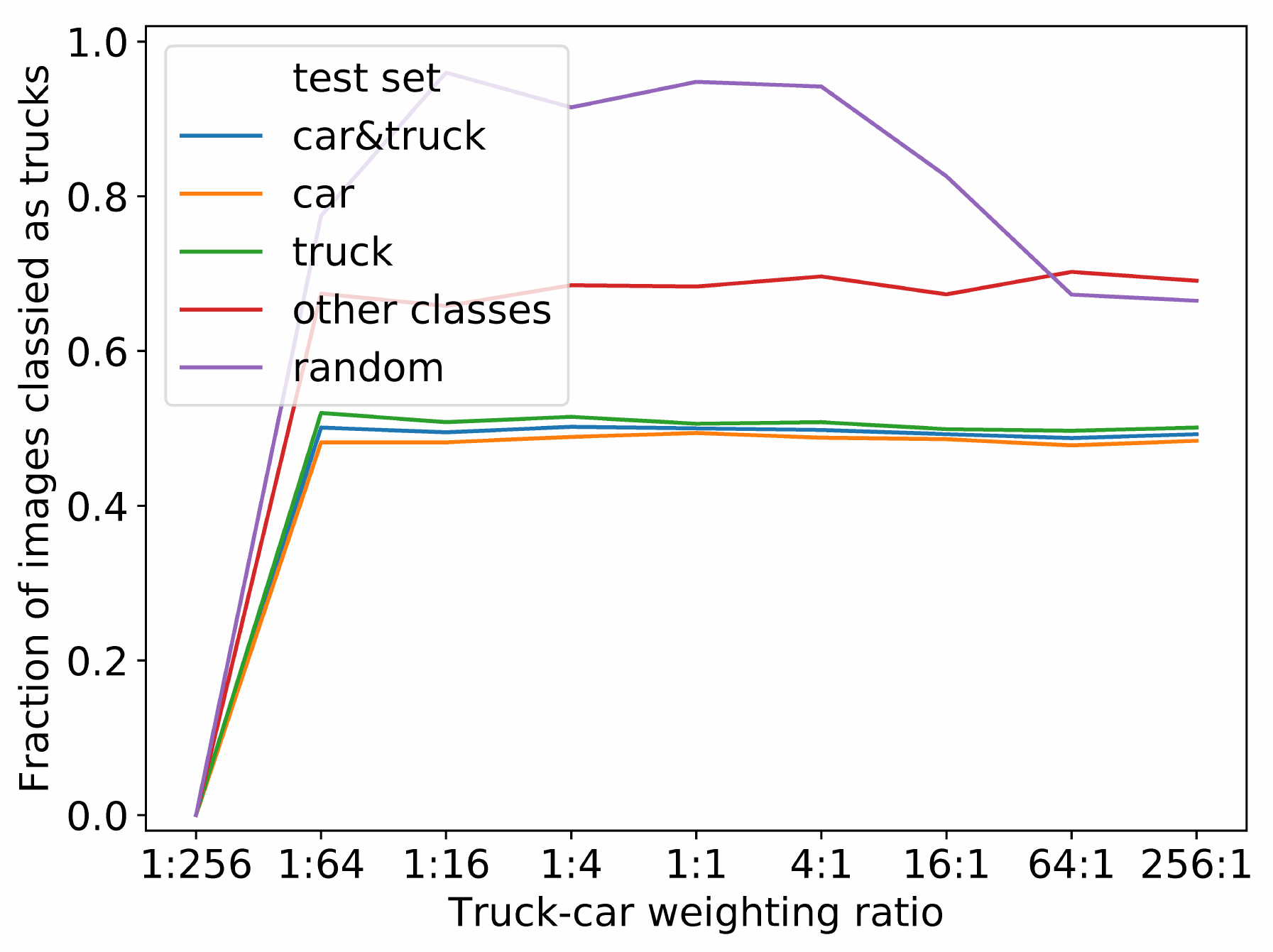}}
    \label{subfig:cifar_iw_cartruck}
    \hfill
    \caption
    {Results for training on automobile and truck classes from CIFAR10. Same setup as ~\ref{fig:cifar}, but without standard deviations over multiple runs.} 
    \label{fig:cartruck}
\end{figure*}

\begin{figure*}[!t]
    \centering
    \subcaptionbox{CIFAR10 cat and dog test images.}%
    [.47\textwidth]{\includegraphics[width=.47\textwidth]{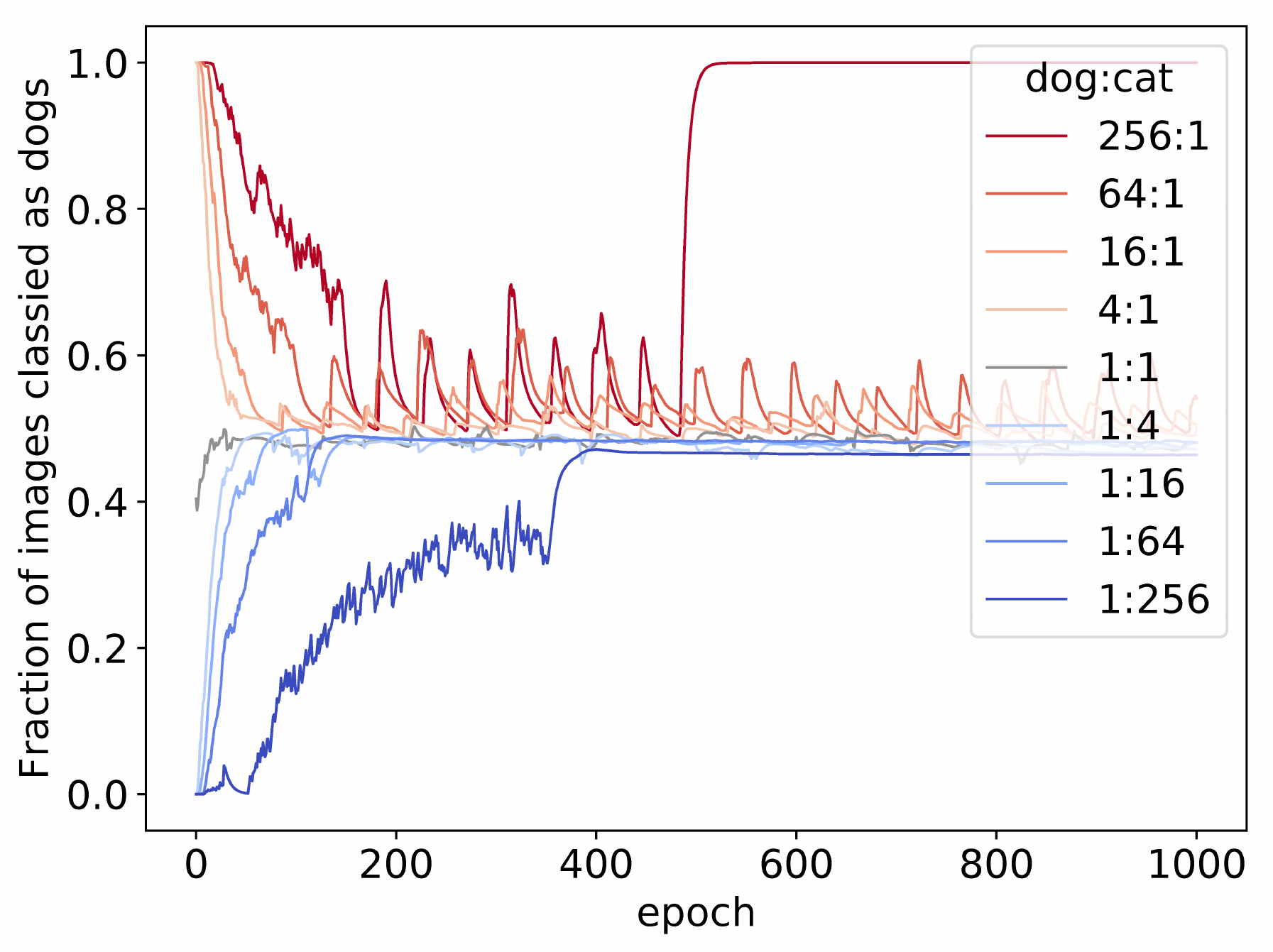}}
    \label{fig:cifar_adam_test}
    \hfill
    \subcaptionbox{CIFAR10 test images from non-cat/dog classes.}%
    [.47\textwidth]{\includegraphics[width=.47\textwidth]{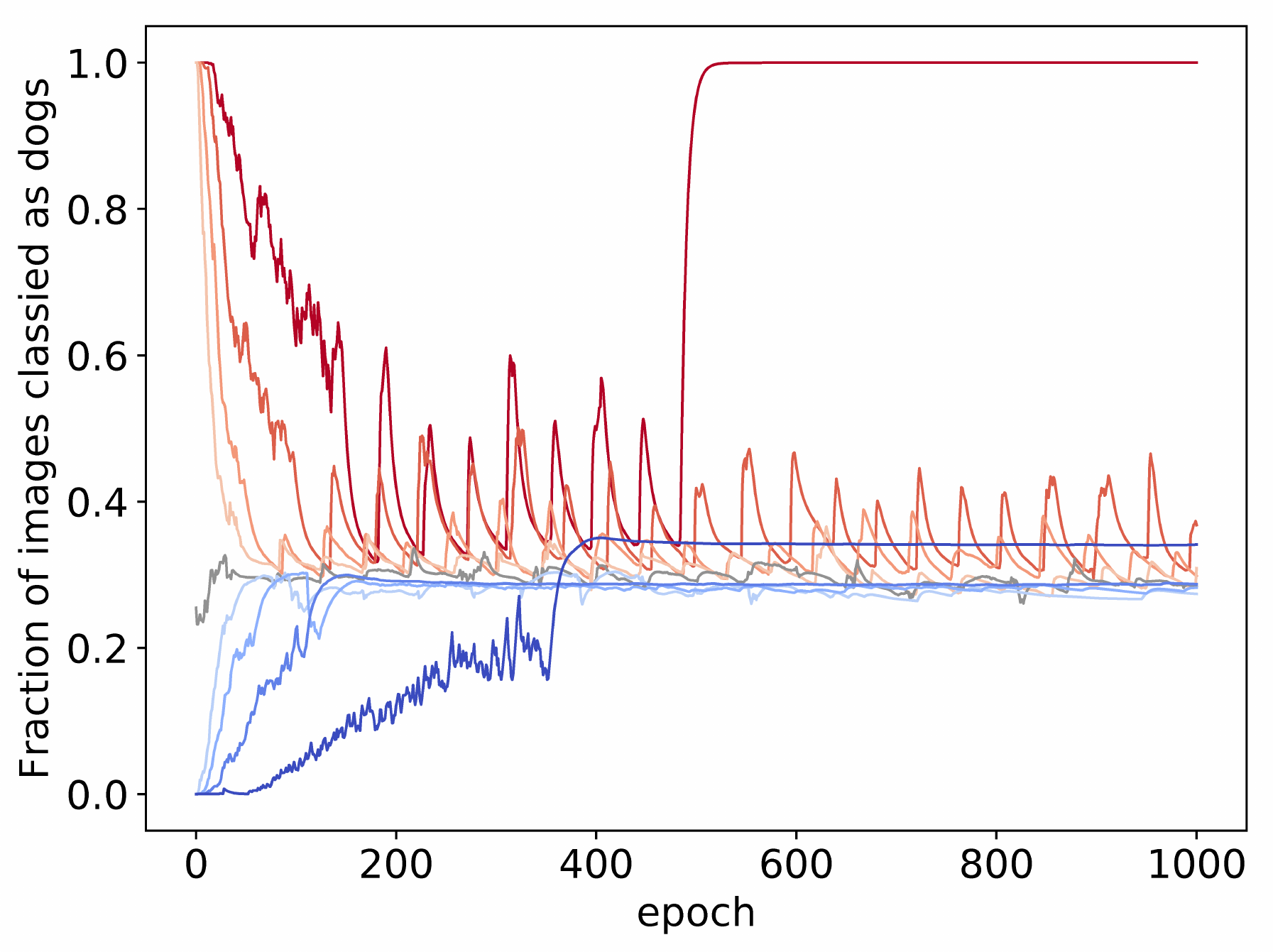}}
    \label{fig:cifar_adam_other}
    \vskip\baselineskip
    \subcaptionbox{Random images.}%
    [.47\textwidth]{\includegraphics[width=.47\textwidth]{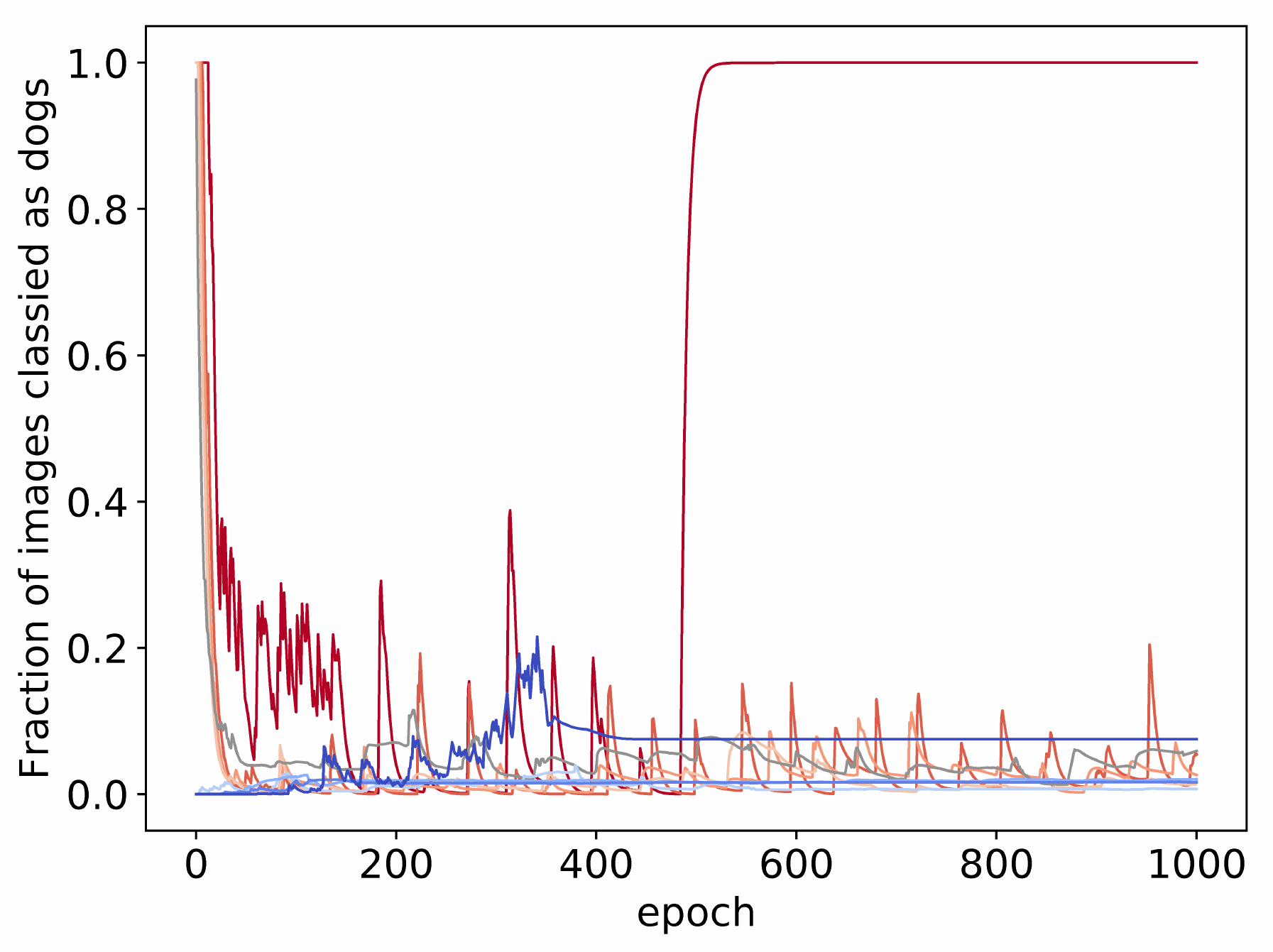}}
    \label{fig:cifar_adam_random}
    \hfill
    \subcaptionbox{Classification ratios after training on cat/dog images.}%
    [.47\textwidth]{\includegraphics[width=.47\textwidth]{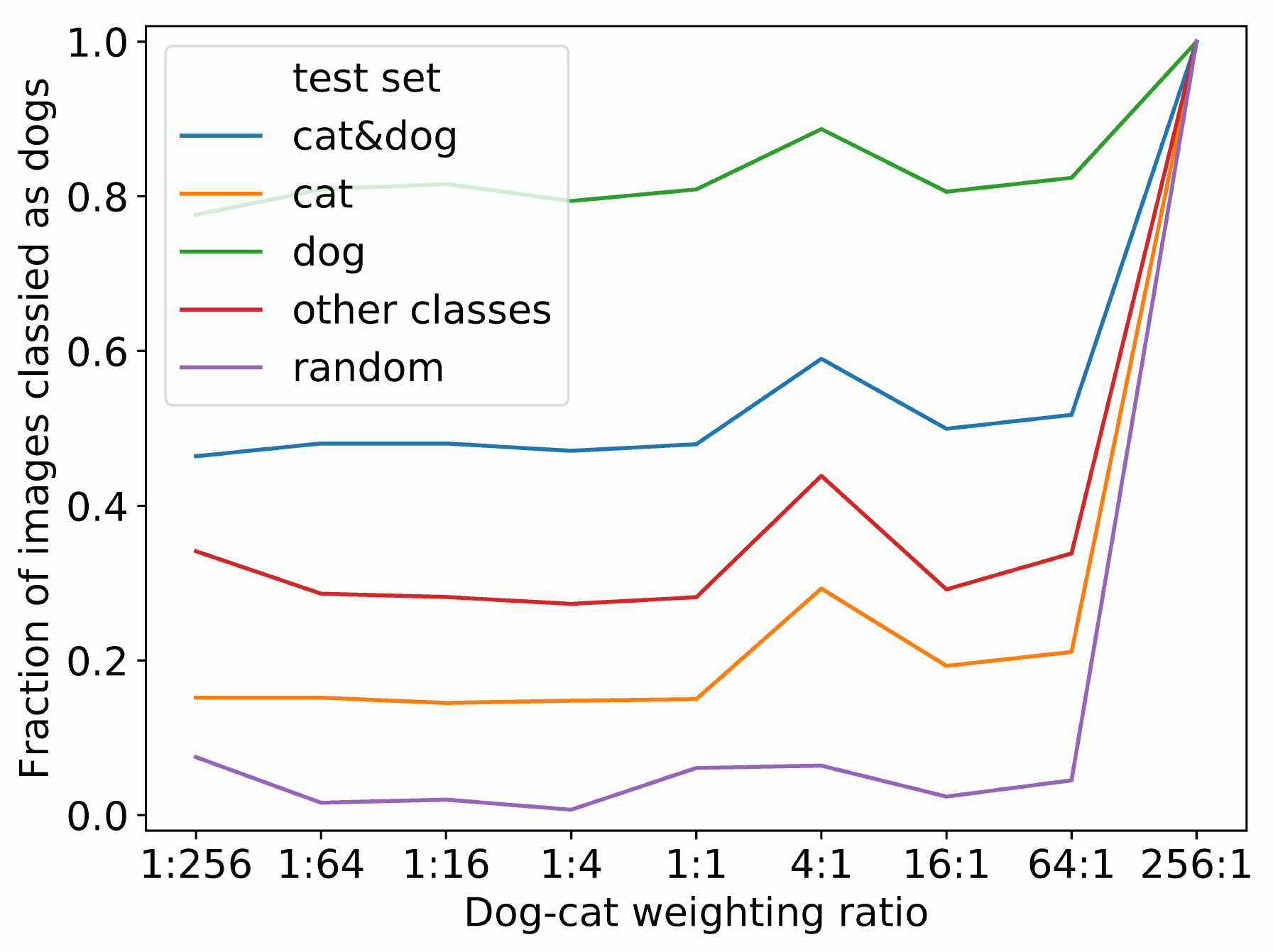}}
    \label{subfig:cifar_iw_adam}
    \hfill
    \caption
    {Results from training a convolutional network with the Adam optimizer with learning rate $1e-4$, $\beta_1=0.9$, $\beta_2=0.999$, $\epsilon=1e-8$. The setup and all other model hyperparameters are the same as in ~\ref{fig:cifar}, but without standard deviations over multiple runs.} 
    \label{fig:adam}
\end{figure*}

\end{document}